\documentclass[10pt,twocolumn,twoside]{IEEEtran}
\usepackage{graphicx}
\usepackage{amsmath}
\usepackage{amssymb}
\usepackage[tight]{subfigure}
\usepackage{cite}
\usepackage{bm}
\usepackage{multirow}
\usepackage{booktabs}
\usepackage{makecell}
\usepackage{algpseudocode}
\usepackage{booktabs}
\usepackage{color}
\usepackage{siunitx}
\usepackage{epstopdf}

\usepackage[pagebackref=true,breaklinks=true,letterpaper=true,colorlinks,bookmarks=false,linkcolor=blue,citecolor=blue]{hyperref}

\begin{document}
\newcommand{\comments}[1]{}
\newcommand{\etal}{\textit{et al}.}
\newcommand{\ie}{\textit{i}.\textit{e}.}
\newcommand{\eg}{\textit{e}.\textit{g}.}
\renewcommand{\algorithmicrequire}{\textbf{Input:}} %
\renewcommand{\algorithmicensure}{\textbf{Output:}} %

\title{Learning Spatial and Spatio-Temporal Pixel Aggregations for Image and Video Denoising}

\author{Xiangyu Xu, \thanks{X. Xu is with Robotics Institute, Carnegie Mellon University, Pittsburgh, PA 15213, USA (email: xuxiangyu2014@gmail.com).} 
	\and Muchen Li, \thanks{M. Li is with University of British Columbia, Canada (email: muchenli1997@gmail.com).} 
	\and Wenxiu Sun, \thanks{W. Sun is with SenseTime Research, Hong Kong, 999077 (email: irene.wenxiu.sun@gmail.com).}
	\and Ming-Hsuan Yang \thanks{M.-H. Yang is with School of Engineering, University of California, Merced, CA 95343, USA (e-mail: mhyang@ucmerced.edu).}
}

%
%

\maketitle

\begin{abstract}
Existing denoising methods typically restore clear results by aggregating pixels from the noisy input.
Instead of relying on hand-crafted aggregation schemes, we propose to explicitly learn this process with deep neural networks.
We present a spatial pixel aggregation network  and learn the pixel sampling and averaging strategies for image denoising. 
The proposed model naturally adapts to image structures and can effectively improve the denoised results.
Furthermore, we develop a spatio-temporal pixel aggregation network for video denoising to efficiently sample pixels across the spatio-temporal space.
Our method is able to solve the misalignment issues caused by large motion in dynamic scenes.
In addition, we introduce a new regularization term for effectively training the proposed video denoising model.
We present extensive analysis of the proposed method and demonstrate that our model performs favorably against the state-of-the-art image and video denoising approaches on both synthetic and real-world data.
\end{abstract}


\IEEEpeerreviewmaketitle

\section{Introduction}
\label{sec:intro}
Image and video capturing systems are often degraded by noise including shot noise of photons and read noise from sensors~\cite{healey1994radiometric}.
This problem is exacerbated for the images and videos captured in low-light scenarios or by cellphone cameras with small-apertures. 
To address the problem, different image denoising algorithms have been proposed for generating high-quality images and video frames from the noisy inputs~\cite{tomasi1998bilateral,non-local-2005,bm3d2007,benchmarking_denoising,vbm3d2007,vbm4d2012,liu2014fast,KPN}.

\begin{figure}[t]
	\footnotesize
	\begin{center}
		\begin{tabular}{c}
\includegraphics[width = 0.9\linewidth]{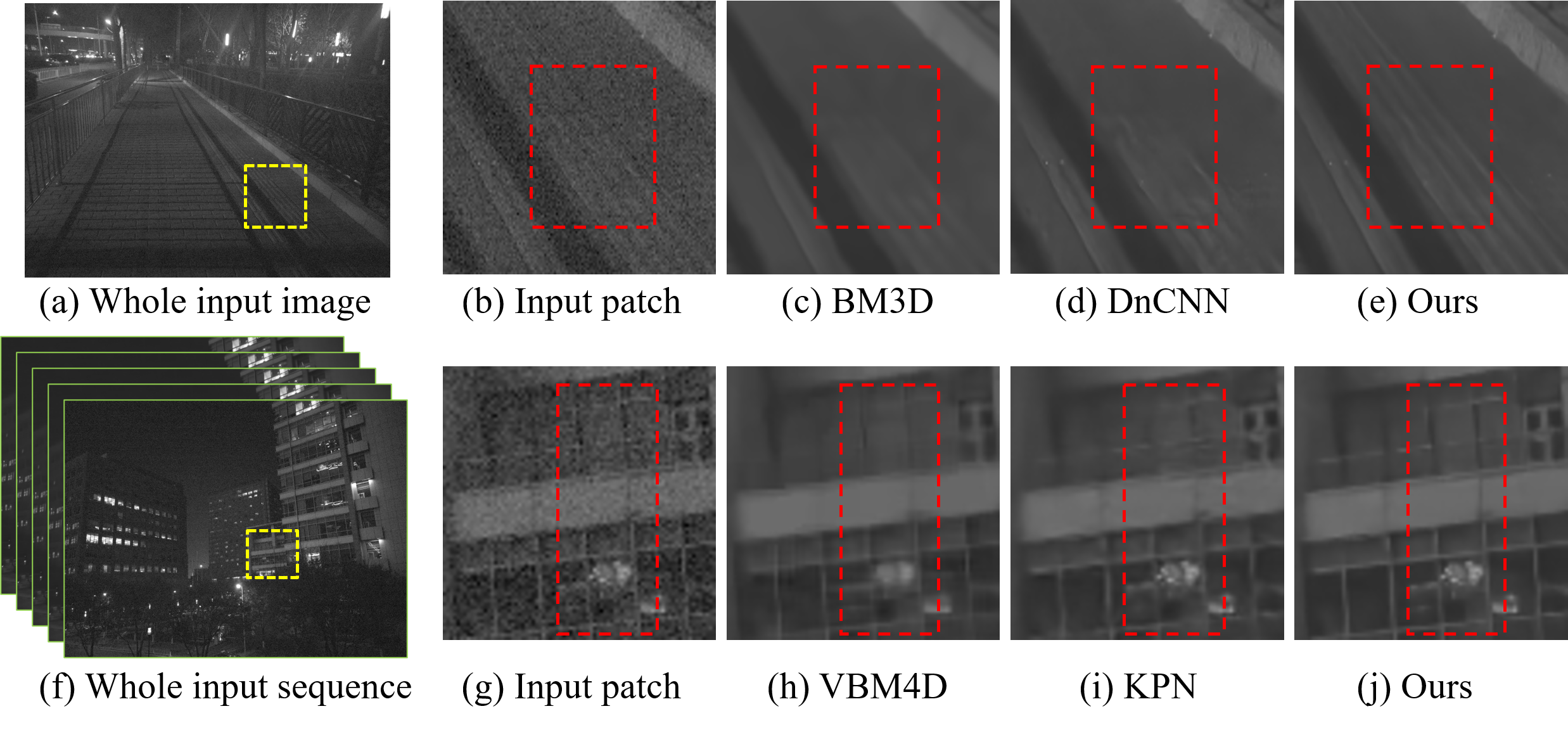} \\
		\end{tabular}
	\end{center}
			\vspace{-2mm}
	\caption{%
		Denoising results on image and video sequence captured by a cellphone.
		Compared to the existing classical (BM3D~\cite{bm3d2007}, VBM4D~\cite{vbm4d2012}) and deep learning based (DnCNN~\cite{dncnn}, KPN~\cite{KPN}) methods, 
		the proposed algorithm achieves better denoising results with fewer artifacts on both single image (top) and multi-frame input (bottom).
		(g) is a cropped region from the reference frame of the input sequence.}
	\label{fig:teaser} 
		\vspace{-5mm}
\end{figure}

The success of most denoising methods stems from the fact that averaging multiple independent observations of the same signal leads to lower variance than the original observations. 
Mathematically, this is formulated as:
\begin{align}
\label{eq:motivation}
	Var(\frac{1}{N}\sum_{i=1}^{N}x_{(i)})=\frac{1}{N}Var(x),
\end{align}
where $Var$ denotes the variance operation. $x$ is a noisy pixel, and $\{x_{(i)}, i=1,2,...,N\}$ are $N$ $i.i.d.$ observations of it. 
Since it is difficult to obtain multiple observations of the same pixel, 
existing denoising algorithms~\cite{non-local-2005,bm3d2007,vbm3d2007,vbm4d2012,tomasi1998bilateral} usually sample similar pixels from the input image and aggregate them with weighted averaging.
The sampling grid $\mathcal{N}$ and averaging weights $\mathcal{F}$ are usually data-dependent and spatially-variant as the distribution of similar pixels depends 
on local image structures.
The strategies to decide $\mathcal{N}$ and $\mathcal{F}$ are the key factors distinguishing different denoising approaches.
As a typical example, 
the bilateral smoothing model~\cite{tomasi1998bilateral} samples pixels in a local square region and computes the weights with Gaussian functions.
In addition, the BM3D~\cite{bm3d2007} and VBM4D~\cite{vbm4d2012} methods search relevant pixels by block matching, and the averaging weights are decided using empirical Wiener filter.
However, these approaches usually use hand-crafted schemes for sampling and weighting pixels, which do not always perform well in complex scenarios as shown in Figure~\ref{fig:teaser}(c) and (h).

Recently,  numerous denoising methods based on deep convolutional neural networks (CNNs)~\cite{dncnn,remez2017deep,claus2019videnn} have been proposed. 
These models exploit a large amount of training data to learn the mapping function from the noisy image to the desired clear output.
However, the CNNs usually use spatially-invariant and data-independent convolution kernels whereas the denoising process is spatially-variant and data-dependent~\cite{bm3d2007}.
Thus, very deep structures are needed for these methods to achieve high non-linearities to implicitly approximate the spatially-variant and data-dependent process, which is not as efficient and concise as the aggregation-based formulation.
In addition, the CNN-based approaches do not explicitly manipulate the input pixels to constrain the output space and may generate corrupted image textures and over-smoothing artifacts as shown in Figure~\ref{fig:teaser}(d).

To address the aforementioned problems, we propose a pixel aggregation network to explicitly integrate the pixel aggregation process with data-driven methods for image denoising. 
Specifically, we use CNNs to estimate a spatial sampling grid $\mathcal{N}$ for each location in the noisy image. 
To aggregate the sampled pixels, we  predict the averaging weights $\mathcal{F}$ for each sample.
Note that both $\mathcal{N}$ and $\mathcal{F}$ are content-aware and can adapt to image structures. 
Finally, the denoised output can be obtained by combining $\mathcal{F}$ and $\mathcal{N}$ with weighted averaging in an end-to-end network.
The advantages of the proposed denoising method are as follows.
First, we improve the pixel aggregation process by learning from data instead of relying on hand-crafted schemes.
Second, compared to other deep learning based methods, the proposed model can better adapt to image structures and preserve details with the spatially-variant and data-dependent sampling and averaging strategies. 
In addition, our algorithm directly filters the noisy input, which thereby constrains the output space.
As shown in Figure~\ref{fig:teaser}(e), the proposed network generates clearer result with fewer artifacts.
Note that while one can simply sample pixels from a rigid grid similar to the kernel prediction network (KPN)~\cite{KPN}, it often leads to limited receptive field and cannot efficiently exploit the structure information in the images. 
Moreover, the irrelevant sampling locations of the rigid sampling scheme may negatively affect the denoising performance.
In contrast, the proposed method can sample pixels in a dynamic manner to better adapt to image structures and increase the receptive field without sampling more locations.

In addition to single images, we can also use the proposed method in video denoising.
A straightforward approach for this is to apply 2D pixel aggregation on each frame separately and then fuse the results by pixel-wise summation.
However, this simple strategy is not effective in handling videos with large motion, where few reliable sampling locations can be found in neighboring regions for pixel aggregation. %
To address this issue, we need to allocate more sampling locations on the frames with higher reliability and discard the frames with drastic motion.
As such, this requires our algorithm to be able to search pixels across the spatio-temporal space of the input video.
Instead of predicting 2D sampling grid and averaging weights,
we develop a spatio-temporal pixel aggregation network for each location in the desired output to adaptively select the most informative pixels in the neighboring video frames.
The proposed spatio-temporal method naturally solves the large motion issues by capturing dependencies between 3D locations and sampling on more reliable frames.
Our method can effectively deal with the misalignment caused by dynamic scenes and reduce the artifacts of existing video denoising approaches~\cite{vbm4d2012,KPN} as shown in Figure~\ref{fig:teaser}(j).

In this paper, we make the following contributions.
First, we exploit the strength of both aggregation-based methods and the deep neural networks, and propose a new algorithm to explicitly learn the pixel aggregation process for image denoising. 
Second, we extend the spatial pixel aggregation to the spatial-temporal domain to better deal with large motion in video denoising, which further reduces artifacts and improves performance. 
In addition, we introduce a regularization term to facilitate training the video denoising model.
Extensive experiments on benchmark datasets demonstrate that our method compares favorably against state-of-the-arts on both single image and video inputs.

\section{Related Work}
We discuss the state-of-the-art denoising algorithms as well as recent methods on learning dynamic operations for image filtering, and put the proposed algorithm in proper context.
\subsection{Image and Video Denoising}
Existing denoising methods are developed based on explicit or implicit pixel aggregations~\cite{gonzalez2002digital,tomasi1998bilateral,non-local-2005,bm3d2007,vbm3d2007,vbm4d2012,liu2014fast}.
Gaussian~\cite{gonzalez2002digital} and bilateral~\cite{tomasi1998bilateral} smoothing models sample pixels from a local window and compute averaging weights using Gaussian functions.
The non-local means (NLM)~\cite{non-local-2005} aggregates image pixels globally and decides the weights with patch similarities.
On the other hand, the BM3D~\cite{bm3d2007} method searches pixels with block matching and uses transform domain collaborative filtering for weighted averaging.

As videos contain temporal information and more pixel observations, the VBM3D~\cite{vbm3d2007} and VBM4D~\cite{vbm4d2012} algorithms extend the BM3D scheme  by grouping more similar patches in the spatio-temporal domain. 
In addition, optical flow has also been exploited in video denoising~\cite{liu2010high, liu2014fast} to aggregate pixels.
However, existing video denoising methods are less effective for videos with fast and complex motion. 
In contrast to the above approaches with hand-crafted strategies for sampling and averaging pixels, our method explicitly learns the pixel aggregation process from data for denoising. 
Furthermore, the proposed spatio-temporal model handles large motion in video denoising without optical flow estimation.

Recently, numerous image and video denoising~\cite{dncnn,remez2017deep,rnn_denoising,godard2017deep,KPN,nlnet,n3net,nlrn,claus2019videnn,zuo2018convolutional,zhang2018ffdnet} methods based on deep learning have been developed. 
In particular, CNNs with residual connections have been used to directly learn a mapping function from the noisy input to the denoised output
\cite{dncnn, remez2017deep, claus2019videnn}.
On the other hand, 
recurrent neural networks (RNNs)~\cite{rnn_denoising, godard2017deep} have also been used for exploiting the temporal structure of videos to learn the mapping function for multiple frame input.
While these networks are effective in removing noise, the adopted activation functions may lead to information loss~\cite{inverting-features}.
In addition, directly synthesizing images with deep neural networks does not enforce constrained output space and thereby tends to generate oversmoothing artifacts.
In contrast, the proposed algorithm can directly manipulate the input frames and adaptively aggregate pixels across the spatial and spatio-temporal space, which effectively addresses the above issues.

\subsection{Burst Denoising}
This work is also related to the burst denoising algorithms~\cite{godard2017deep,KPN,kokkinos2019iterative,liu2014fast,hasinoff2016burst} in that they rely on the same basic idea of averaging multiple independent observations as in \eqref{eq:motivation}.
Specifically, burst denoising exploits multiple short frames captured in a burst to approximate multiple independent observations of the same pixel. 
As a typical example, Hasinoff \etal~\cite{hasinoff2016burst} propose a computational photography pipeline to merge a burst of frames to reduce noise and increase dynamic range.
Recently, Kokkinos \etal~\cite{kokkinos2019iterative} use deep learning to solve this problem and propose iterative residual CNNs for further improving the performance.
While these methods have achieved impressive results for image denoising, they usually assume small motion between different frames in the burst and thus do not always work well for videos. 
To solve this problem, Mildenhall~\etal~\cite{KPN} propose to predict convolution kernels for burst denoising which can also work well for video input.
However, these kernels use rigid sampling grids which cannot exploit local image structures well. 
Furthermore, these kernels are less effective in handling large misalignment caused by camera shakes or dynamic scenes.
In contrast, we propose spatially-variant and data-dependent sampling grids as well as a spatio-temporal model to address these issues. 
\subsection{Learning Dynamic Filtering}
\label{sec:flexible}
In recent years, deep neural networks have been used to learn dynamic operations for image filtering~\cite{jia2016dynamic,stn,sttn,dai2017deformable}.
In~\cite{jia2016dynamic}, a dynamic network is proposed 
to learn filtering weights for video and stereo prediction.
Similar approaches have been developed for video interpolation~\cite{niklaus2017video} and view synthesis~\cite{park2017transformation}.
However, these methods only consider pixels from a fixed region, which often leads to limited receptive field and can be easily affected by irrelevant sampling locations.
On the other hand, Jaderberg \etal~\cite{stn} develop spatial transformer networks~\cite{stn} for more flexible feature extractions in image classification.
While this scheme enables data-dependent sampling strategies, 
the filtering process is still spatially-invariant as it only learns a global affine transformation.
In contrast, Dai \etal~\cite{dai2017deformable} propose spatial deformable kernels for object detection, which considers local geometric transformations.
As this method uses fixed convolution weights for different input images, it is only effective for high-level vision tasks. 
The approaches using rigid weights are likely to generate oversmoothing artifacts in image restoration similar to those based on Gaussian filters.
While \cite{zhu2019deformable} learns both the convolution locations and weights, it explains the learned weights as modulation for adjusting the signal amplitude of the input features, and thereby is not suitable for the denoising problem.
In addition, these methods~\cite{dai2017deformable,zhu2019deformable} cannot sample pixels from the spatio-temporal space, and thus does not perform well for video inputs.

The proposed pixel aggregation network can be seen as a novel dynamic operation for image filtering.
Our model learns both the data-dependent and spatially-variant sampling and weighting schemes, 
and thus solves the problems of the aforementioned algorithms.
More importantly, our method enables adaptive sampling in the spatio-temporal space for effective video processing.

\begin{figure}[t]
	\footnotesize
	\begin{center}
		\begin{tabular}{c}
			\includegraphics[width = 0.99\linewidth]{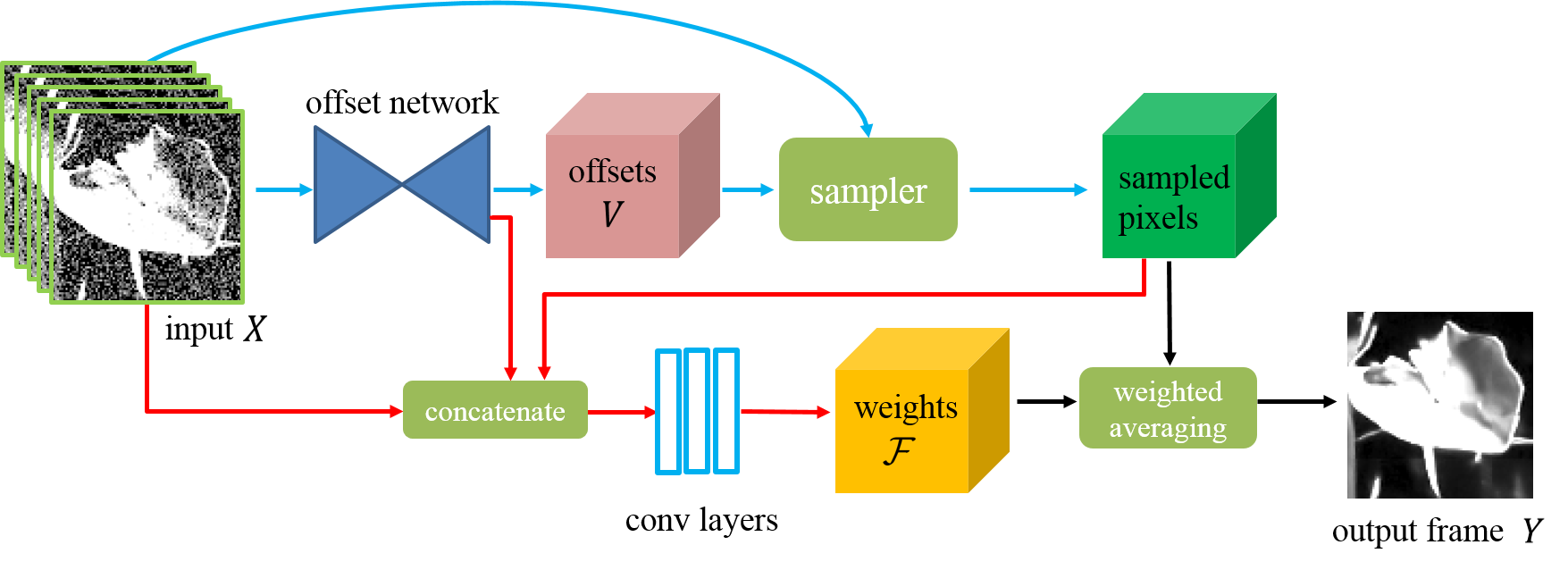} \\
		\end{tabular}
	\end{center}
	\vspace{-3mm}
	\caption{
		Overview of the proposed algorithm.
		We first learn a deep CNN (\ie~the offset network) to estimate the offsets $V$ of the sampling grid.
		We then sample pixels from the noisy input $X$ according to the predicted offsets.
		Furthermore, we concatenate the sampled pixels, the input and the features of the offset network to estimate the averaging weights $\mathcal{F}$.
		Finally, we can generate the denoised output frame $Y$ by aggregating the sampled pixels with the learned weights $\mathcal{F}$. 
		Note that the proposed system can deal with both single image and video sequence inputs.
	}
	\label{fig:pipeline}
	\vspace{-4mm}
\end{figure}
\section{Proposed Algorithm}
We propose a neural network to learn pixel aggregations for image and video denoising.
Compared to most CNN-based denoising methods based on 
data-independent and spatially-invariant kernels, 
the proposed model can better adapt to image structures and preserve details with data-dependent and spatially-variant sampling as well as averaging strategies.
Specifically, we use a neural network to predict the sampling locations $\mathcal{N}$ and averaging weights $\mathcal{F}$ for each pixel of the noisy input.
These two components are integrated for both spatial and spatio-temporal pixel aggregation. 
Instead of directly regressing the spatial coordinates of $\mathcal{N}$, we learn the offsets $V$ for a rigid sampling grid and deform the rigid grid accordingly.

An overview of the proposed algorithm is shown in Figure~\ref{fig:pipeline}.
We first train a deep CNN for estimating the offsets of the sampling grid. 
Next, we sample pixels from the noisy input according to the predicted offsets, and estimate the weights by concatenating the sampled pixels, the noisy input and the features of the offset network.
Finally, we generate the denoised output by averaging the sampled pixels with the learned weights. 

\begin{table*}[t]
	\footnotesize
	\begin{center}
		\caption{Number of feature maps for each layer of our network. We show the structure of the offset network in Figure~\ref{fig:3d_ker}(b). The ``conv layers" are presented in Figure~\ref{fig:pipeline}. ``Ck" represents the k-th convolution layer in each part of our model. $n$ is the number of the sampled pixels of the adaptive sampling grid.}
		\label{table: net_details}
		\begin{tabular}{c|c|c|c|c|c|c|c|c|c|c|c|c}
			\hline 
			\multirow{2}*{Layer name} & \multicolumn{10}{c|}{offset network} & \multicolumn{2}{c}{conv layers} \\
			\cline{2-13}
			       & C1-3 & C4-6 & C7-9 & C10-12 & C13-15 & C16-18 & C19-21 & C22-24 & C25-26 & C27 & C1-2 & C3 \\
			       \hline
			       number of feature maps & 64 & 128 & 256 & 512 & 512 & 512 & 256 & 128 & 128 & $n\times$3 & 64 & $n$ \\
			\hline
		\end{tabular}
	\end{center}
\vspace{-5mm}
\end{table*}

\subsection{Learning to Aggregate Pixels}
\label{sec:3d kernel}
For a noisy image $X \in \mathbb{R}^{h\times w}$ where $h$ and $w$ represent the height and width,
the spatial pixel aggregation  for denoising can be formulated as:
	\begin{align}
	\label{eq:filtering}
	Y(u,v)=\sum_{i=1}^{n}  X(u+u_i,v+v_i) \mathcal{F}(u,v,i),
	\end{align}
where $(u,v)$ is a pixel on the denoised output $Y \in \mathbb{R}^{h\times w}$. 
In addition, $\mathcal{N}(u,v)=\{(u_i,v_i)|i=1,2,\dots,n\}$ represents the sampling grid with $n$ sampling locations, and $\mathcal{F}\in \mathbb{R}^{h\times w \times n}$ represents the weights for averaging pixels.
For example,
\begin{align}
\{(\hat{u}_i,\hat{v}_i)\}=\{(-1,-1),\dots,(0,0),\dots,(1,1)\}, 
\end{align}
defines a rigid sampling grid with $n=9$ and size $3\times3$.

In the proposed pixel aggregation network (PAN), the adaptive sampling grid can be generated
by combining the predicted offsets $V \in \mathbb{R}^{h\times w \times n \times 2}$ and the rigid grid:
	\begin{align}
	u_i=\hat{u}_i+V(u,v,i,1), \label{eq:V1}\\
	v_i=\hat{v}_i+V(u,v,i,2). \label{eq:V2}
	\end{align}
Note that both $u_i$ and $v_i$ are functions of $(u,v)$, which indicates that our denoising process is spatially-variant.
Since the offsets in $V$ are usually fractional, we use bilinear interpolation to sample the pixels $X(u+{u}_i,v+{v}_i)$ in a way similar to \cite{super-slowmo}.

\begin{figure}[t]
	\footnotesize
	\begin{center}
		\begin{tabular}{c}
			\includegraphics[width = 0.75\linewidth]{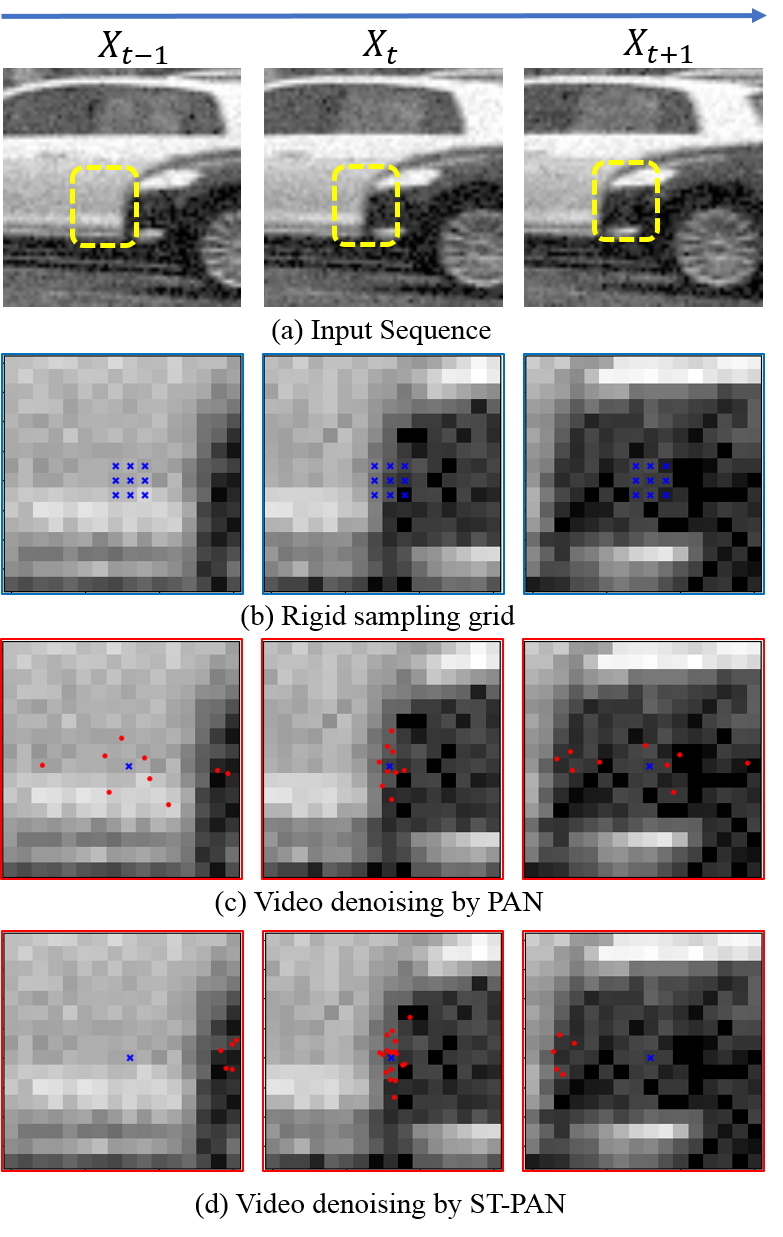} \\
		\end{tabular}
	\end{center}
	\vspace{-2mm}
	\caption{{Illustration of the video denoising process of the ST-PAN model. (a) a noisy video sequence $\{X_{t-1},X_{t},X_{t+1}\}$.
			The patches in the following rows are cropped from the yellow box in the corresponding frames. 
			The center blue point of patch $X_t$ in (b)-(d) indicates the reference pixel to be denoised.
			(b) The rigid sampling method has limited receptive field and cannot exploit the structure information. 
			Furthermore, it does not handle misalignment issues. 
			(c) The proposed PAN model can adapt to image structures in $X_t$ and increase the receptive field without sampling more pixels.
			However, it does not perform well on large motion where there are few reliable pixels available in frame $X_{t-1}$ and $X_{t+1}$.
			(d) The proposed ST-PAN model aggregates pixels across the spatial-temporal space, and distributes more sampling locations on more reliable frames.
		}
	}
	\label{fig:introduction}
	\vspace{-3mm}
\end{figure}

After the adaptive sampling, we can recover the clear output $Y$ by combining the sampled pixels with the learned weights $\mathcal{F}$ as in \eqref{eq:filtering}.
The weights of $\mathcal{F}$ are also spatially-variant and content-aware, which is different from the typical CNNs with fixed uniform convolution kernels.
Note that while we can simply use a rigid sampling grid (Figure~\ref{fig:introduction}(b)) and only learn the averaging weights,
it often leads to a limited receptive field and cannot efficiently exploit the structure information in the images.
Furthermore, irrelevant sampling locations in the rigid grid may negatively affect the denoising results.
In contrast, our adaptive grid naturally adapts to the image structures and increases the receptive field without sampling more pixels.

\begin{figure}[t]
	\footnotesize
	\begin{center}
		\begin{tabular}{c}
			\hspace{0mm}
			\includegraphics[width = 0.9\linewidth]{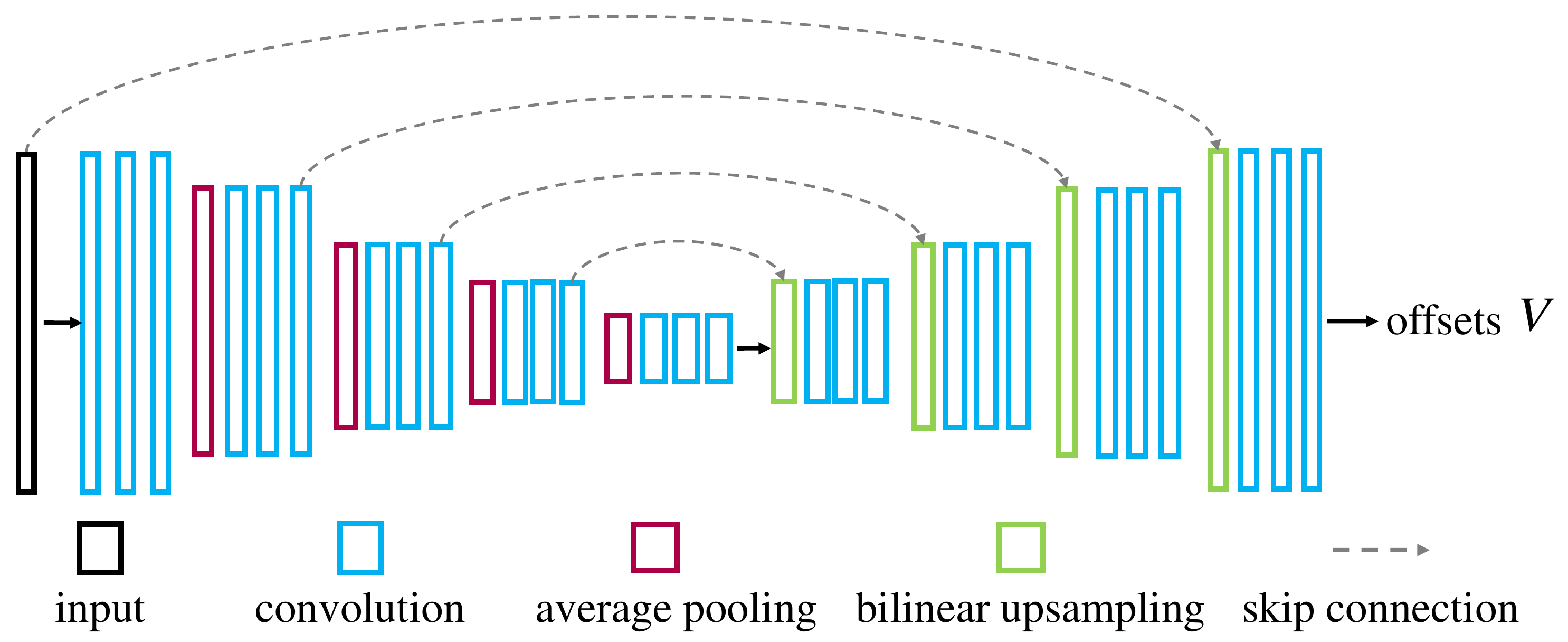} \\
		\end{tabular}
	\end{center}
	\vspace{-2mm}
	\caption{Architecture of the offset network. A U-Net with skip links is used to fuse low-level and high-level features for estimating the offsets.}
	\label{fig:3d_ker}
	\vspace{-5mm}
\end{figure}
\vspace{1ex}
{\flushleft\bf Spatio-temporal pixel aggregation.} 
The proposed method can be easily extended for video denoising. 
Suppose that we have a noisy video sequence $\{X_{t-\tau},\dots,X_t,\dots,X_{t+\tau}\}$, where $X_t$ is the reference frame.
A straightforward approach to process this input is to apply the PAN model to each frame separately and then fuse the outputs with weighted sum, as shown in Figure~\ref{fig:introduction}(c).
However, this simple 2D strategy is not effective in handling videos with large motion, where few reliable pixels can be found in the regions of neighboring frames (\eg~the center regions of frame $X_{t-1}$ and $X_{t+1}$ in Figure~\ref{fig:introduction}).
To address this issue, we need to distribute more sampling locations on the frames with higher reliability (\eg~the reference frame $X_{t}$) and avoid the frames with severe motion.
An effective solution should be able to search pixels across the spatial-temporal space of the input videos.

In this work, 
we develop a spatio-temporal pixel aggregation network (ST-PAN) for video denoising which adaptively selects the most informative pixels in the spatio-temporal space. 
The ST-PAN directly takes the concatenated video frames $X \in \mathbb{R}^{h\times w \times (2\tau+1)}$ as input,
and the denoising process can be formulated as:
\begin{equation}
\label{eq:3d_filtering}
Y(u,v,t)=\sum_{i=1}^{n} X(u+u_i,v+v_i,t+t_i) \mathcal{F}(u,v,i),
\end{equation}
where $t+t_i$ denotes the sampling coordinate in the temporal dimension, and $n$ is the number of pixels of the 3D sampling grid. 
Similar to \eqref{eq:V1}-\eqref{eq:V2},
we generate the sampling grid  by predicting 3D offsets $V \in \mathbb{R}^{h\times w \times n \times 3}$.
To sample pixels across the video frames, we introduce the trilinear interpolation 
in which $X(u+u_i,v+v_i,t+t_i)$ can be computed by:
\begin{align}
\sum_{p=1}^h \sum_{q=1}^w \sum_{j=t-\tau}^{t+\tau} X(p,q,j)  \cdot \max(0,1-|u+u_i-p|) \nonumber\\ 
\cdot \max(0,1-|v+v_i-q|)  \cdot \max(0,1-|t+t_i-j|),
\end{align}
where only the pixels closest to $(u+u_i,v+v_i,t+t_i)$ in the 3D space of $X$ contribute to the interpolated result.
Since the trilinear sampling mechanism is differentiable, we can learn the ST-PAN model 
in an end-to-end manner.
The proposed ST-PAN model naturally solves the large motion issues by capturing dependencies between 3D locations and sampling on more reliable frames, as illustrated in Figure~\ref{fig:introduction}(d).
Furthermore,
our method can effectively deal with the misalignment caused by dynamic scenes and reduce cluttered boundaries and ghosting artifacts generated by existing video denoising approaches~\cite{vbm4d2012,KPN} as shown in Section~\ref{sec: eval_our_data}.

{\flushleft\bf Gamma correction.}
As the noise is nonlinear in the sRGB space~\cite{anderson1996proposal,ie1999iec}, we train the denoising model in the linear raw space. 
With the linear output $Y$, we conduct Gamma correction to generate the final result for better perceptual quality:
	\begin{align}
	\phi(Y) & = {\begin{cases}
		12.92 Y, &Y\leq 0.0031308, \\
		(1+\alpha)Y^{1/2.4}-\alpha, & Y > 0.0031308, 
		\end{cases}}%
	\end{align}
where $\phi$ is the sRGB transformation function for Gamma correction, and $\alpha =0.055$. The hyper-parameters of $\phi$ are directly obtained from \cite{ie1999iec}, and more detailed explanations can be found in~\cite{anderson1996proposal,ie1999iec}.

\subsection{Network Architecture}
\label{sec:network}
The offset network in Figure~\ref{fig:pipeline} takes a single frame as input for image denoising, and a sequence of $2\tau+1$ neighboring frames for video denoising.
As shown in Figure~\ref{fig:3d_ker}(b), we adopt a U-Net architecture~\cite{ronneberger2015u} which has been widely used in pixel-wise estimation tasks~\cite{learning_to_see_in_the_dark,xu2018rendering}. 
The U-Net is an encoder-decoder network where the encoder sequentially transforms the input frames into lower-resolution feature embeddings, and the decoder correspondingly expands the features back to full resolution estimates.
We perform pixel-wise summation with skip connections between the same-resolution layers in the encoder and decoder to jointly use low-level and high-level features for the estimation task.
Since the predicted weights are to be applied to the sampled pixels, it is beneficial to feed these pixels to the weight estimation branch such that the weights can better adapt to the sampled pixels.  
Thus, we concatenate the sampled pixels, noisy input and features from the last layer of the offset network, and feed them to three convolution layers to estimate the averaging weights (Figure~\ref{fig:pipeline}).

All convolution layers use $3\times 3$ kernels with stride $1$.
The feature map number for each layer of our network is shown in Table~\ref{table: net_details}.
We use ReLU~\cite{nair2010rectified} as the activation function for the convolution layers except for the last one which is followed by a Tanh function to output normalized offsets.
As the proposed estimation network is fully convolutional, it is able to handle arbitrary spatial size during inference.

\subsection{Loss Function}
With the predicted result $Y$ and ground truth image $Y_{gt}$ in  the linear space, we can use an $L_1$ loss to train our network for single image denoising:
\begin{align}
l(Y,Y_{gt})=\|\phi(Y)-\phi(Y_{gt})\|_1,
\end{align}
where Gamma correction is performed to emphasize errors in darker regions and generate more perceptually pleasant results.
We do not use $L_2$ loss in this work as it often leads to oversmoothing artifacts~\cite{lim2017enhanced,zhang2018residual}.

\vspace{1ex}
{\noindent \bf{Regularization term for video denoising.}} %
Since the ST-PAN model samples pixels across the video frames, 
it is possible that the training process is stuck to local minimum 
where all the sample locations only lie around the reference frame.
To alleviate this problem and encourage the network to exploit more temporal information, 
we introduce a regularization term to have subsets of the sampled pixels individually learn the 3D aggregation process. %

We split the $N$ sampling locations in the spatio-temporal grid $\{(u_n,v_n,t_n)\}$ into $s$ groups: $\mathcal{N}_1,...,\mathcal{N}_s$, and each group consists of $N/s$ points.
Similar to~\eqref{eq:3d_filtering}, the filtered result of the $i$-th pixel group can be computed by::
\begin{align}
Y_i(u,v,t) = s \sum_{j\in \mathcal{N}_i} X(u+u_j,v+v_j,t+t_j)F(u,v,j),
\end{align}
where $i\in \{1,2,...,s\}$, and the multiplier $s$ is used to match the scale of $Y$.
With $Y_i$ for regularization, the final loss function for video denoising is:
\begin{align}
\label{eq:regu}
l(Y,Y_{gt})+\eta \gamma^{m} \sum_{i=1}^{s} l(Y_i,Y_{gt}).
\end{align}
The regularization process for each $Y_i$ is slowly reduced during training, where the hyperparameters $\eta$ and $\gamma$ are used to control the annealing process. 
$m$ is the iteration number.
At the beginning of the network optimization, $\eta \gamma^{m}\gg1$ and the second term is prominent, which encourages the network to find the most informative pixels for each subset of the sampled pixels.
This constraint is lessened as $m$ increases, and the whole sampling grid learns to rearrange the sampling locations such that all the pixel groups, \ie~different parts of the learned pixel aggregation model, can perform collaboratively.
Note that the temporal consistency \cite{ren2018deep,davy2019non,dvd} is implicitly enforced in this per-frame loss function as the ground truth image $Y_{gt}$ changes smoothly across the sequence.
\begin{figure*}[t]
	\footnotesize
	\begin{center}
		\begin{tabular}{cccccc}
			\includegraphics[height = 0.09\linewidth]{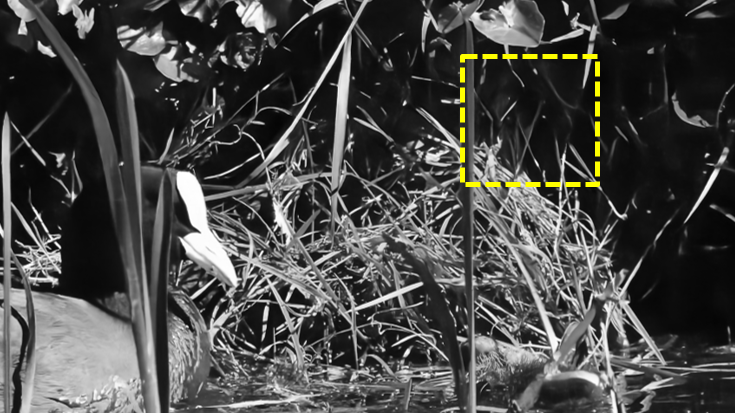} & \hspace{-4mm}
			\includegraphics[height = 0.09\linewidth]{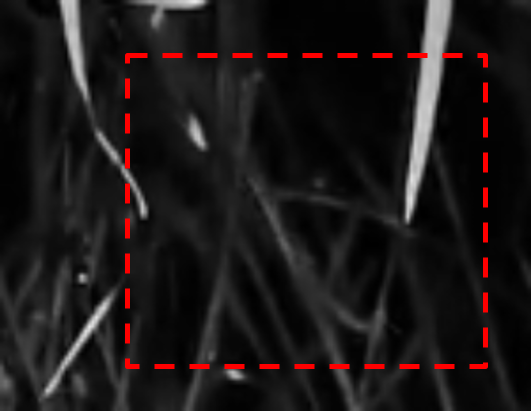}  & \hspace{-4mm}
			\includegraphics[height = 0.09\linewidth]{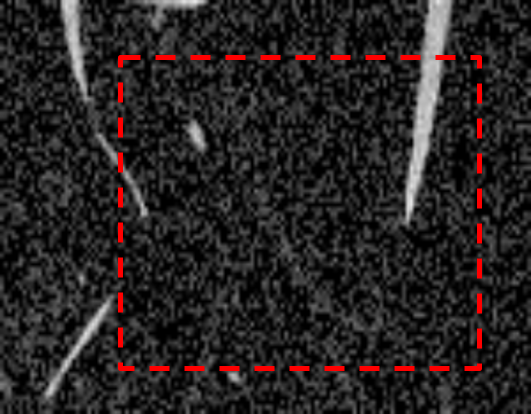} & \hspace{-4mm}
			\includegraphics[height = 0.09\linewidth]{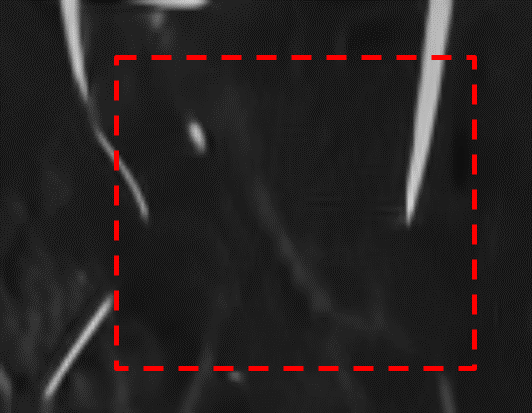} & \hspace{-4mm}
			\includegraphics[height = 0.09\linewidth]{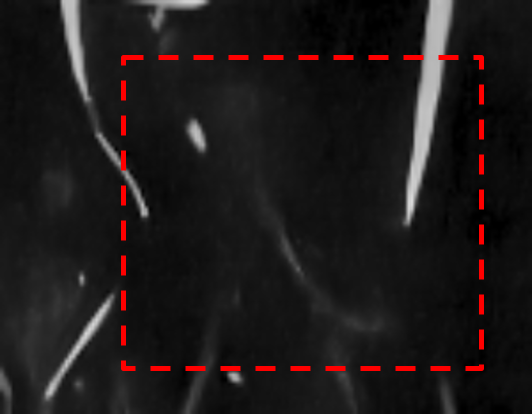} & \hspace{-4mm}
			\includegraphics[height = 0.09\linewidth]{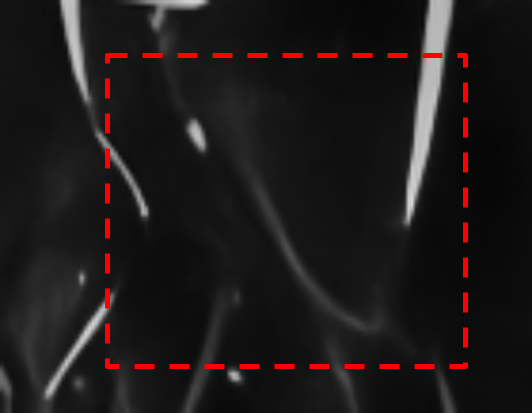} \\
			\includegraphics[height = 0.09\linewidth]{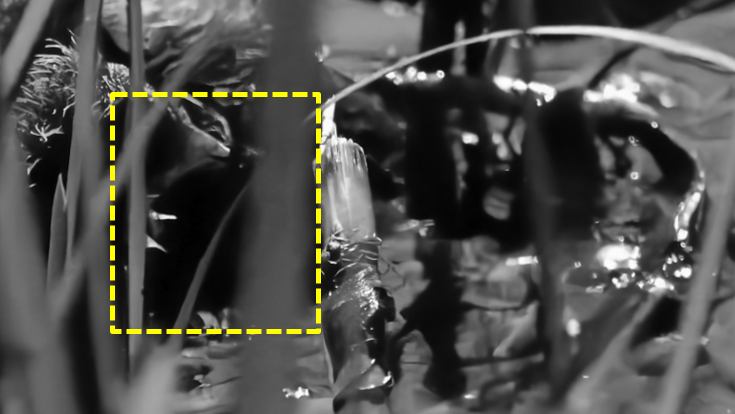} & \hspace{-4mm}
			\includegraphics[height = 0.09\linewidth]{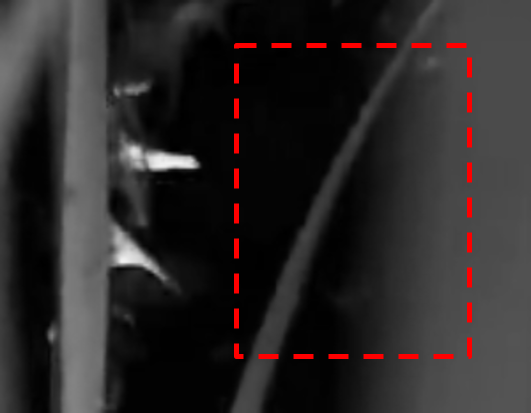}  & \hspace{-4mm}
			\includegraphics[height = 0.09\linewidth]{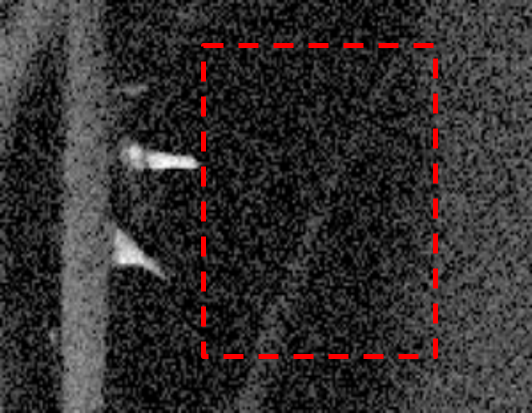} & \hspace{-4mm}
			\includegraphics[height = 0.09\linewidth]{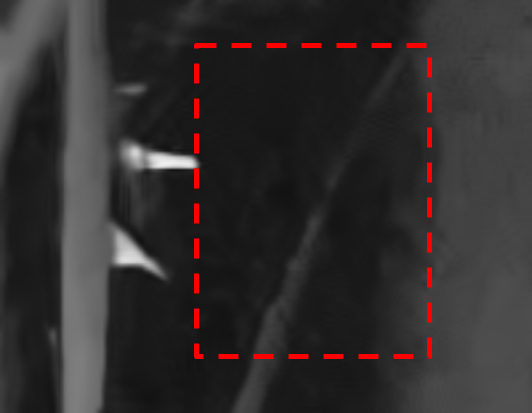} & \hspace{-4mm}
			\includegraphics[height = 0.09\linewidth]{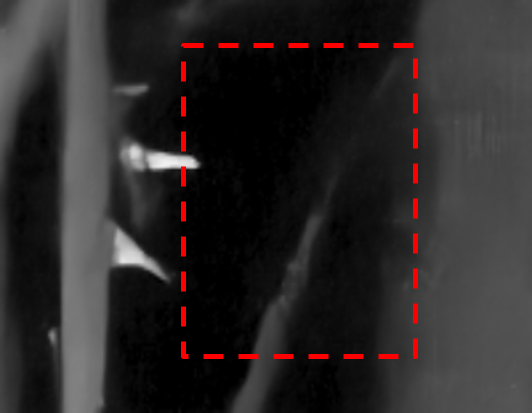} & \hspace{-4mm}
			\includegraphics[height = 0.09\linewidth]{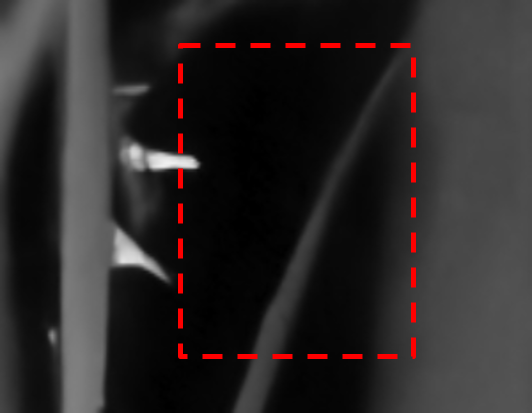} \\
			(a) Whole output & \hspace{-4mm} (b) Ground truth & \hspace{-4mm} (c) Input & \hspace{-4mm} (d) BM3D~\cite{bm3d2007} & \hspace{-4mm} (e) DnCNN~\cite{dncnn} & \hspace{-4mm} (f) PAN \\
			\includegraphics[height = 0.09\linewidth]{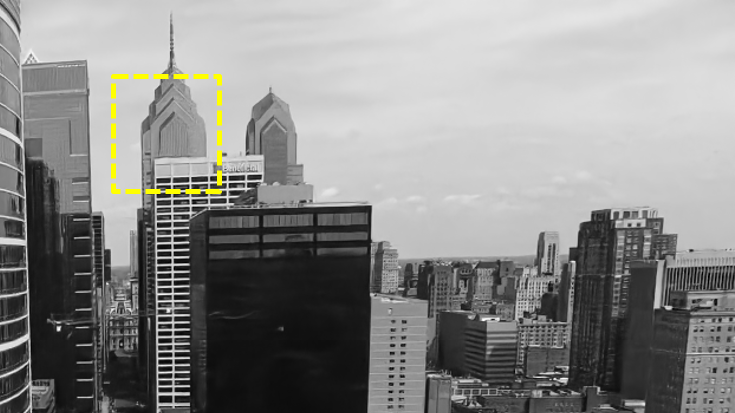} & \hspace{-4mm}
			\includegraphics[height = 0.09\linewidth]{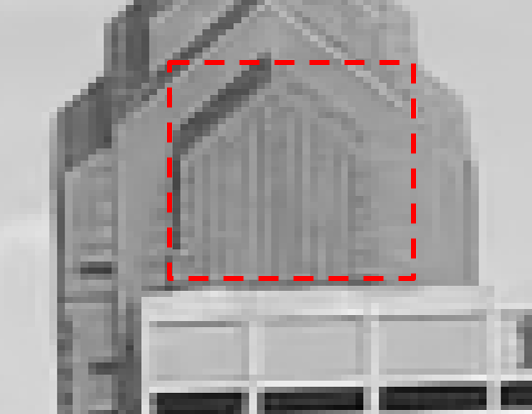}  & \hspace{-4mm}
			\includegraphics[height = 0.09\linewidth]{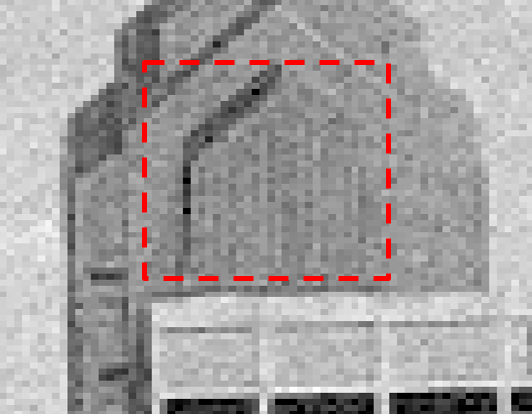} & \hspace{-4mm}
			\includegraphics[height = 0.09\linewidth]{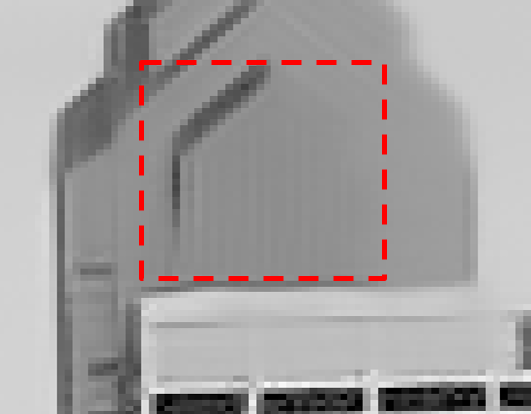} & \hspace{-4mm}
			\includegraphics[height = 0.09\linewidth]{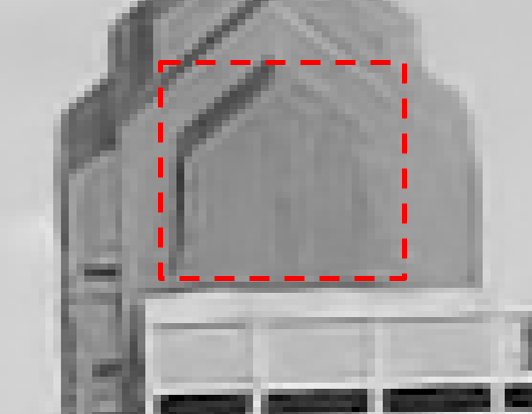} & \hspace{-4mm}
			\includegraphics[height = 0.09\linewidth]{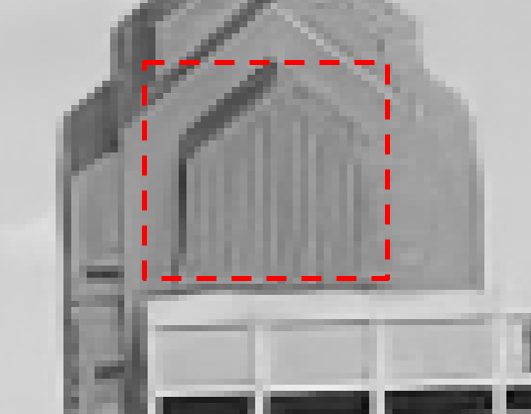} \\
			\includegraphics[height = 0.09\linewidth]{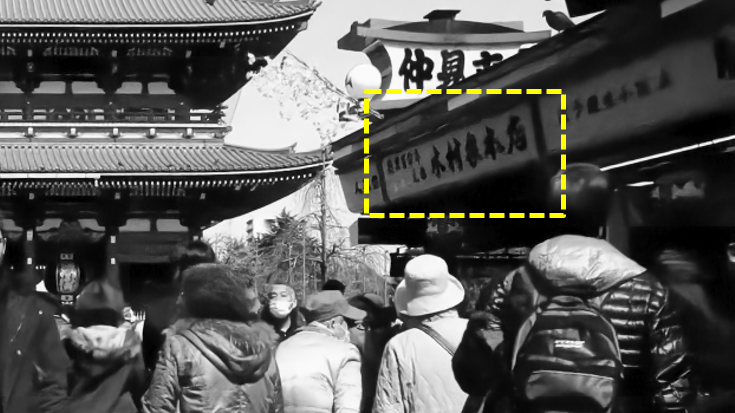} & \hspace{-4mm}
			\includegraphics[height = 0.09\linewidth]{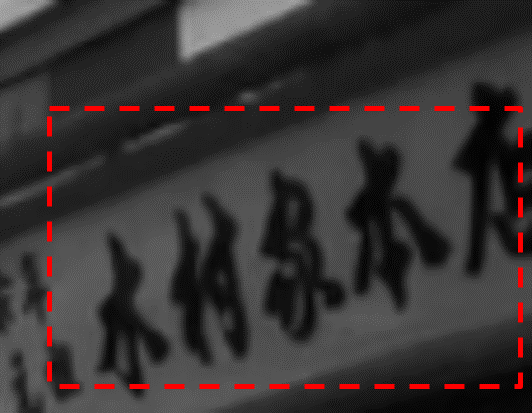}  & \hspace{-4mm}
			\includegraphics[height = 0.09\linewidth]{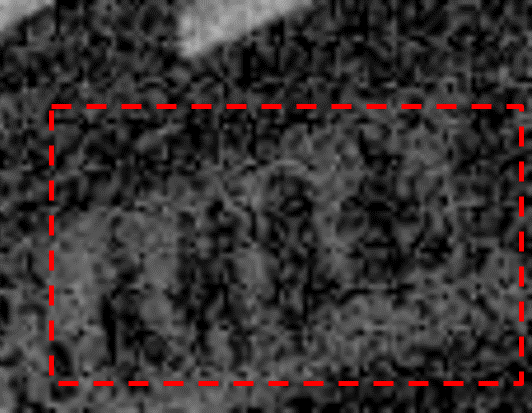} & \hspace{-4mm}
			\includegraphics[height = 0.09\linewidth]{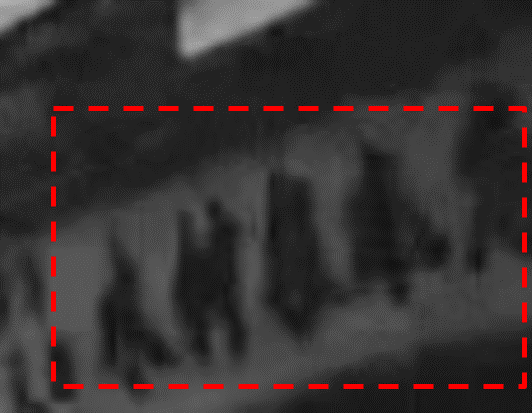} & \hspace{-4mm}
			\includegraphics[height = 0.09\linewidth]{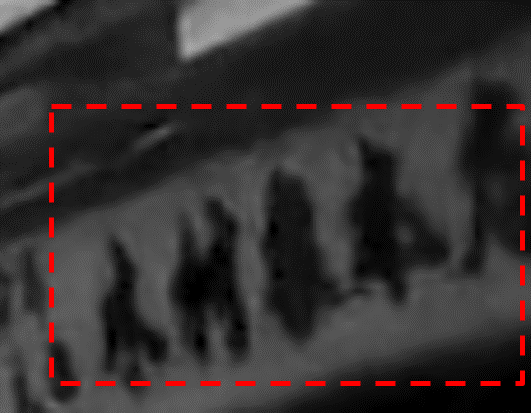} & \hspace{-4mm}
			\includegraphics[height = 0.09\linewidth]{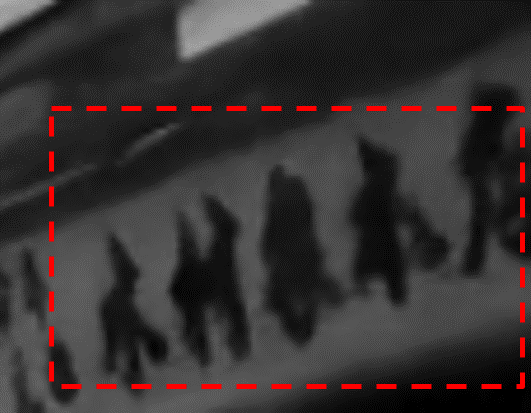} \\
			(g) Whole output & \hspace{-4mm} (h) Ground truth & \hspace{-4mm} (i) Input & \hspace{-4mm} (j) VBM4D~\cite{vbm4d2012} & \hspace{-4mm} (k) KPN~\cite{KPN} & \hspace{-4mm} (l) ST-PAN \\
		\end{tabular}
	\end{center}
	\vspace{-1mm}
	\caption{Results from the synthetic dataset for single image (first row) and video denoising (second row). The proposed method generates clearer results with fewer artifacts.}
	\vspace{-2mm}
	\label{fig:synthetic_video_1}
\end{figure*}
\begin{table*}[]
	\centering
	\footnotesize
	\caption{Quantitative evaluation of single image denoising on the synthetic dataset. \#1-4 are the 4 testing subsets. ``LOW" and ``HIGH" represent different noise levels, which respectively correspond to $\sigma_s=2.5\times10^{-3},\sigma_r=10^{-2}$ and $\sigma_s=6.4\times10^{-3},\sigma_r=2\times10^{-2}$.
		{\color{red}Red} and {\color{blue}blue} indicate the first and second best performance for each noise level.
		\label{table:synthetic_result_image}}
	\begin{tabular}{c|l|c|c|c|c|c|c|c|c|c|c}
		\hline
		\multicolumn{1}{l|}{}                   &                              & \multicolumn{2}{c|}{\#1}                                     & \multicolumn{2}{c|}{\#2}                                     & \multicolumn{2}{c|}{\#3}                                     & \multicolumn{2}{c|}{\#4}                                     & \multicolumn{2}{c}{Average}                                 \\ \cline{3-12} 
		\multicolumn{1}{l|}{\multirow{-2}{*}{Noise}} & \multirow{-2}{*}{Algorithms} & PSNR                         & SSIM                          & PSNR                         & SSIM                          & PSNR                         & SSIM                          & PSNR                         & SSIM                          & PSNR                         & SSIM                          \\ \hline \hline
		& Reference frame              & 26.75                        & 0.6891                        & 28.08                        & 0.7333                        & 27.37                        & 0.5843                        & 27.96                        & 0.7064                        & 27.54                        & 0.6782                        \\ %
		& NLM~\cite{non-local-2005}                          & 31.04                        & 0.8838                        & 31.51                        & 0.9025                        & 33.35                        & 0.8687                        & 31.71                        & 0.8663                        & 31.90                        & 0.8803                        \\ %
		& BM3D~\cite{bm3d2007}                         & 33.00                        & 0.9196                        & 32.63                        & 0.9245                        & 35.16                        & 0.9172                        & 33.09                        & 0.9028                        & 33.47                        & 0.9160                        \\ %
		& DnCNN~\cite{dncnn}                        & 35.30                        & 0.9499                        & 34.54                        & 0.9498                        & 37.45                        & 0.9436                        & 36.22                        & 0.9494                        & 35.88                        & 0.9482                        \\ %
		& KPN~\cite{KPN}                          & 35.23                        & 0.9526                        & 34.38                        & 0.9493                        & 37.50                        & 0.9451                        & 36.18                        & 0.9526                        & 35.82                        & 0.9499                        \\ %
		& PAN                          & {\color{red} 35.40} & {\color{red} 0.9535} & {\color{red} 34.57} & {\color{blue} 0.9507} & {\color{red} 37.64} & {\color{red} 0.9465} & {\color{red} 36.41} & {\color{red} 0.9538} & {\color{red} 36.01} & {\color{red} 0.9511} \\ \cline{2-12} 
		& KPN~\cite{KPN}, $\sigma$ blind         & 35.18                        & 0.9492                        & 34.20                        & 0.9484                        & 37.39                        & 0.9438                        & 36.05                        & 0.9508                        & 35.71                        & 0.9480                        \\ %
		& DnCNN~\cite{dncnn}, $\sigma$ blind       & 35.19                        & 0.9500                        & 34.38                        & 0.9479                        & 37.28                        & 0.9417                        & 36.06                        & 0.9491                        & 35.73                        & 0.9472                        \\ %
		\multirow{-9}{*}{LOW}                    & PAN, $\sigma$ blind                    & \color{blue} 35.33                        & \color{blue} 0.9531                        & \color{blue} 34.55                        & \color{red} 0.9508                        &\color{blue} 37.57                        & \color{blue} 0.9458                        & \color{blue} 36.35                        & \color{red} 0.9538                        & \color{blue} 35.95                        & \color{blue} 0.9509                        \\ \hline \hline
		& Reference frame              & 22.83                        & 0.5403                        & 23.94                        & 0.5730                        & 23.00                        & 0.3746                        & 23.97                        & 0.5598                        & 23.43                        & 0.5119                        \\ %
		& NLM~\cite{non-local-2005}                          & 28.21                        & 0.8236                        & 28.57                        & 0.8443                        & 30.62                        & 0.8076                        & 28.73                        & 0.8040                        & 29.03                        & 0.8199                        \\ %
		& BM3D~\cite{bm3d2007}                         & 29.96                        & 0.8793                        & 29.81                        & 0.8836                        & 32.30                        & 0.8766                        & 30.27                        & 0.8609                        & 30.59                        & 0.8751                        \\ %
		& DnCNN~\cite{dncnn}                        & 32.30                        & 0.9163                        & 31.54                        & 0.9124                        & 34.55                        & 0.9048                        & 33.26                        & 0.9148                        & 32.91                        & 0.9121                        \\ %
		& KPN~\cite{KPN}                          & 32.32                        & 0.9198                        & 31.44                        & 0.9120                        & 34.74                        & 0.9085                        & 33.28                        & 0.9200                        & 32.94                        & 0.9151                        \\ %
		& PAN                          & {\color{red} 32.49} & {\color{red} 0.9226} & {\color{red} 31.62} & {\color{red} 0.9153} & {\color{red} 34.89} & {\color{red} 0.9121} & {\color{red} 33.51} & {\color{red} 0.9232} & {\color{red} 33.13} & {\color{red} 0.9183} \\ \cline{2-12} 
		& KPN~\cite{KPN}, $\sigma$ blind                    & 32.23                        & 0.9182                        & 31.37                        & 0.9107                        & 34.63                        & 0.9073                        & 33.17                        & 0.9183                        & 32.85                        & 0.9136                        \\ %
		& DnCNN~\cite{dncnn}, $\sigma$ blind                  & 32.19                        & 0.9158                        & 31.42                        & 0.9105                        & 34.40                        & 0.9023                        & 33.08                        & 0.9135                        & 32.77                        & 0.9105                        \\ %
		\multirow{-9}{*}{HIGH}                   & PAN, $\sigma$ blind                    & \color{blue} 32.44                        &\color{blue} 0.9224                        & \color{red} 31.62                        & \color{blue} 0.9152                        & \color{blue} 34.81                        & \color{blue} 0.9109                        & \color{blue}33.46                        & \color{blue}0.9215                        & \color{blue}33.08                        & \color{blue}0.9175                        \\ \hline
	\end{tabular}
\end{table*}

\begin{table*}[]
	\centering
	\footnotesize
	\caption{Quantitative evaluation of video denoising on the synthetic dataset. 
		\#1-4 are the 4 testing subsets.
		``PAN-sep" represents the simple 2D strategy of using PAN for video input.
		``LOW" and ``HIGH" denote different noise levels, which respectively correspond to $\sigma_s=2.5\times10^{-3},\sigma_r=10^{-2}$ and $\sigma_s=6.4\times10^{-3},\sigma_r=2\times10^{-2}$.
		{\color{red}Red} and {\color{blue}blue} indicate the first and second best performance for each noise level.
		\label{table:synthetic_result_video}}
	\begin{tabular}{c|l|c|c|c|c|c|c|c|c|c|c}
		\hline
		\multicolumn{1}{l|}{}                   &                              & \multicolumn{2}{c|}{\#1}                                     & \multicolumn{2}{c|}{\#2}                                     & \multicolumn{2}{c|}{\#3}                                     & \multicolumn{2}{c|}{\#4}                                     & \multicolumn{2}{c}{Average}                                 \\ \cline{3-12} 
		\multicolumn{1}{l|}{\multirow{-2}{*}{Noise}} & \multirow{-2}{*}{Algorithms} & PSNR                         & SSIM                          & PSNR                         & SSIM                          & PSNR                         & SSIM                          & PSNR                         & SSIM                          & PSNR                         & SSIM                          \\ \hline \hline
		& Direct average              & 22.75                        & 0.6880                        & 25.70                        & 0.7777                        & 25.15                        & 0.6701                        & 23.47                        & 0.6842                        & 25.27                        & 0.7050                        \\ %
		& VBM4D~\cite{vbm4d2012}                         & 33.26                        & 0.9326                        & 34.00                        & 0.9469                        & 35.83 & 0.9347& 34.01 & 0.9327 &34.27 & 0.9367                       \\ %
		& KPN~\cite{KPN}                        & 35.61 & \color{blue}0.9597&35.25 & 0.9637&38.18 & 0.9529&36.45 & 0.9604&36.37 & 0.9592                   \\ %
		& PAN-sep                          & 35.66 & 0.9576&{\color{red}35.82} & \color{blue}0.9656&38.19 & 0.9518&{\color{blue}36.80} & 0.9609&{\color{blue}36.62} & 0.9590                      \\ %
		& ST-PAN                          & {\color{red}36.02} & {\color{red}0.9618}& {\color{blue}35.80} & \color{red}0.9666& {\color{red}38.78} & \color{red}0.9580& {\color{red}37.04} & \color{red}0.9624& {\color{red}36.91} & \color{red}0.9622 \\ \cline{2-12} 
		& KPN~\cite{KPN}, $\sigma$ blind         & 35.44 & 0.9577&35.03 & 0.9605&38.03 & 0.9506&36.30 & 0.9586&36.20 & 0.9569                      \\ %
		\multirow{-9}{*}{LOW}                    & ST-PAN, $\sigma$ blind                    & {\color{blue}35.70} & 0.9590&35.47 & 0.9633&{\color{blue}38.35} & \color{blue}0.9538&36.67 & \color{blue} 0.9615&36.55 & \color{blue}0.9594                        \\ \hline \hline
		& Direct average              & 21.96                        & 0.6071                        & 24.78                        & 0.6934                        & 24.34                        & 0.5466                        & 22.81                        & 0.6055                        & 23.47                        & 0.6132                        \\ %
		& VBM4D~\cite{vbm4d2012}                         & 30.34 & 0.8894 & 31.28 & 0.9089&32.66 & 0.8881& 31.33 & 0.8925
		&31.40 & 0.8947                    \\ %
		& KPN~\cite{KPN}                        & 32.92 & \color{blue}0.9344&32.56 & 0.9358&35.59 & 0.9223&33.80 & 0.9355&33.72 & \color{blue}0.9320                  \\ %
		& PAN-sep                          & 32.94 & 0.9309&{\color{red}33.09} & \color{blue}0.9380&35.59 & 0.9208&{\color{blue}34.15} & \color{blue}0.9365&{\color{blue}33.94} & 0.9315                     \\ %
		& ST-PAN                          & {\color{red}33.29} & \color{red}0.9372& {\color{blue}33.05} & \color{red}0.9400& {\color{red}36.17} & \color{red}0.9301& {\color{red}34.40} & \color{red}0.9390& {\color{red}34.23} & \color{red}0.9366 \\ \cline{2-12} 
		& KPN~\cite{KPN}, $\sigma$ blind                    & 32.73 & 0.9302&32.36 & 0.9312&35.39 & 0.9185&33.61 & 0.9309&33.52 & 0.9277                 \\ %
		\multirow{-9}{*}{HIGH}                   & ST-PAN, $\sigma$ blind                    & {\color{blue}33.02} & 0.9327&32.79 & 0.9348&{\color{blue}35.78} & \color{blue}0.9239&34.09 & 0.9361&33.92 & 0.9319                   \\ \hline
	\end{tabular}
\end{table*}

\section{Experimental Results}
We first describe the datasets and implementation details, and then evaluate the proposed algorithm for image and video denoising quantitatively and qualitatively. 
\subsection{Datasets}
For video denoising, we collect 27 high-quality long videos from the Internet,
where each has a resolution of $1080\times1920$ or $720\times1280$  
pixels and a frame rate of $20$, $25$, or \SI{30}{fps}.
We use 23 long videos for training and the other 4 for testing,
which are split into 205 and 65 non-overlapped scenes, respectively.
With the videos containing different scenes, we extract 20K sequences for training where each sequence consists of $2\tau+1$ consecutive frames.
Our test dataset is composed of 4 subsets where each has approximately 30 sequences sampled from the 4 testing videos.
There is no overlap between training and testing videos.
In addition, we use the center frame of each sequence from the video datasets for both training and testing in single image denoising.

Similar to~\cite{KPN}, we generate the noisy input for our models by 
performing inverse Gamma correction and  
adding signal-dependent Gaussian noise, $ \mathcal{N}(0, \sigma_s q+\sigma_r^2)$,
where $q$ represents the intensity of the pixel,
and the noise parameters $\sigma_s$ and $\sigma_r$ are randomly sampled from  $[10^{-4},10^{-2}]$ and $[10^{-3},10^{-1.5}]$, respectively.
When dealing with homoscedastic Gaussian noise, we set the shot noise as $0$ and use the target noise level for the read noise during training similar to \cite{dncnn}.  
In our experiments,
we train the networks in both blind and non-blind manners. 
For the non-blind model, the parameters $\sigma_s$ and $\sigma_r$ are assumed to be known, and the noise level is fed into the network as an additional channel of the input.
Similar to~\cite{KPN},
we estimate the noise level as: $\sqrt{\sigma_r^2 + \sigma_s q_{ref} }$,
where $q_{ref}$ represents the intensity value of the reference frame $X_t$ in video denoising or the input image in single frame denoising.
Similar to \cite{KPN}, we use grayscale inputs for fair comparisons with other methods~\cite{bm3d2007,vbm4d2012,dncnn,KPN}.
\subsection{Training and Parameter Settings}
We learn sampling grids with size $5\times5$ for single image denoising.
For video input, we use size $3\times3\times3$ for the spatio-temporal pixel aggregation to 
reduce GPU memory requirement.
We set $\eta$ and $\gamma$ as $100$ and $0.9998$, respectively.
In addition, we set $s=3$ for the regularization term by default.
During training, we use the Adam optimizer~\cite{kingma2014adam} with the initial learning rate of $2\times10^{-4}$.
We decrease the learning rate by a factor of $0.999991$ per epoch, until it reaches $1\times10^{-4}$.
The batch size is set to be $32$.
We randomly crop $128\times128$ patches from the original input for training the single image model. 
In video denoising, we crop at the same place of all the input frames and set $\tau=2$, such that each training sample has a size of $128\times128\times5$.
In our experiments, we multiply the output of the offset network by $128$ where we assume the largest spatial offset of the sampling grid is smaller than the size of the training patch.
We train the denoising networks for $2\times10^5$ iterations, and the process takes about 50 hours.

\begin{table*}[t]
	\begin{center}
		\scriptsize
		\caption{Quantitative evaluation of single image denoising on homoscedastic Gaussian noise. 
			We directly obtain the PSNRs and SSIMs of the baselines from the original papers,
			and indicate the results that are not available with ``-''.
		}
		\vspace{-2mm}
		\begin{tabular}{ l |c|c|c|c|c|c|c|c|c }
			\hline 
			\multicolumn{1}{l|}{{\multirow{2}*{}}}
			& \multicolumn{1}{c|}{{\multirow{2}*{$\sigma$}}} & \multicolumn{1}{c|}{NLNet~\cite{lefkimmiatis2017non}} & \multicolumn{1}{c|}{N3Net~\cite{Ploetz:2018:NNN}} & \multicolumn{1}{c|}{DnCNN~\cite{dncnn}} & \multicolumn{1}{c|}{NLRN~\cite{liu2018non}} & \multicolumn{1}{c|}{SGN~\cite{gu2019self}} &
			\multicolumn{1}{c|}{DDFN-x5W~\cite{chen2019real}} & \multicolumn{1}{c|}{FOCNet~\cite{jia2019focnet}} & \multicolumn{1}{c}{Ours} \\ 
			\cline{3-10}
			& & PSNR / SSIM & PSNR / SSIM & PSNR / SSIM & PSNR / SSIM &  PSNR / SSIM & PSNR / SSIM &  PSNR / SSIM &  PSNR / SSIM \\
			\hline
			\multirow{3}*{Set12}
			& 15 & - / - & - / - & 32.86 / 0.9031 & 33.16 / 0.9070 &32.85 / 0.9031&32.98 / 0.9052& 33.07 / - & {\bf 33.24} / \bf 0.9110\\
			& 25 & 30.31 / - & 30.55 / - & 30.44 / 0.8622 & 30.80 / 0.8689 &30.41 / 0.8639&30.60 / 0.8668&30.73 / -& \bf 30.97 / \bf 0.8735\\
			& 50 & 27.04 / - & 27.43 / - & 27.18 / 0.7829 & 27.64 / 0.7980 &26.77 / 0.7784&27.46 / 0.7960&27.68 / -& \bf 27.84 / \bf 0.8047\\
			\hline
			\multirow{3}*{BSD68}
			& 15 & 31.52 / - & - / -	& 31.73 / 0.8907 & 31.88 / 0.8932 &31.67 / 0.8897&31.83 / 0.8935& 31.83 / - & \bf 31.91 / \bf 0.8980\\
			& 25 & 29.03 / - & 29.30 / - & 29.23 / 0.8278 & 29.41 / 0.8331 &29.03 / 0.8251&29.35 / 0.8331&29.38 / -& \bf 29.52 / \bf 0.8410\\
			& 50 & 26.07 / - & 26.39 / - & 26.23 / 0.7189 & 26.47 / 0.7298 &25.42 / 0.7020&26.42 / 0.7302&26.50 / - & \bf 26.63 / \bf 0.7433\\
			\hline
		\end{tabular}
		\label{tab:more_img}
		\vspace{-2mm}
	\end{center}
\end{table*}

\begin{figure}[t]
	\footnotesize
	\begin{center}
		\begin{tabular}{c}
			\includegraphics[width = 0.95\linewidth]{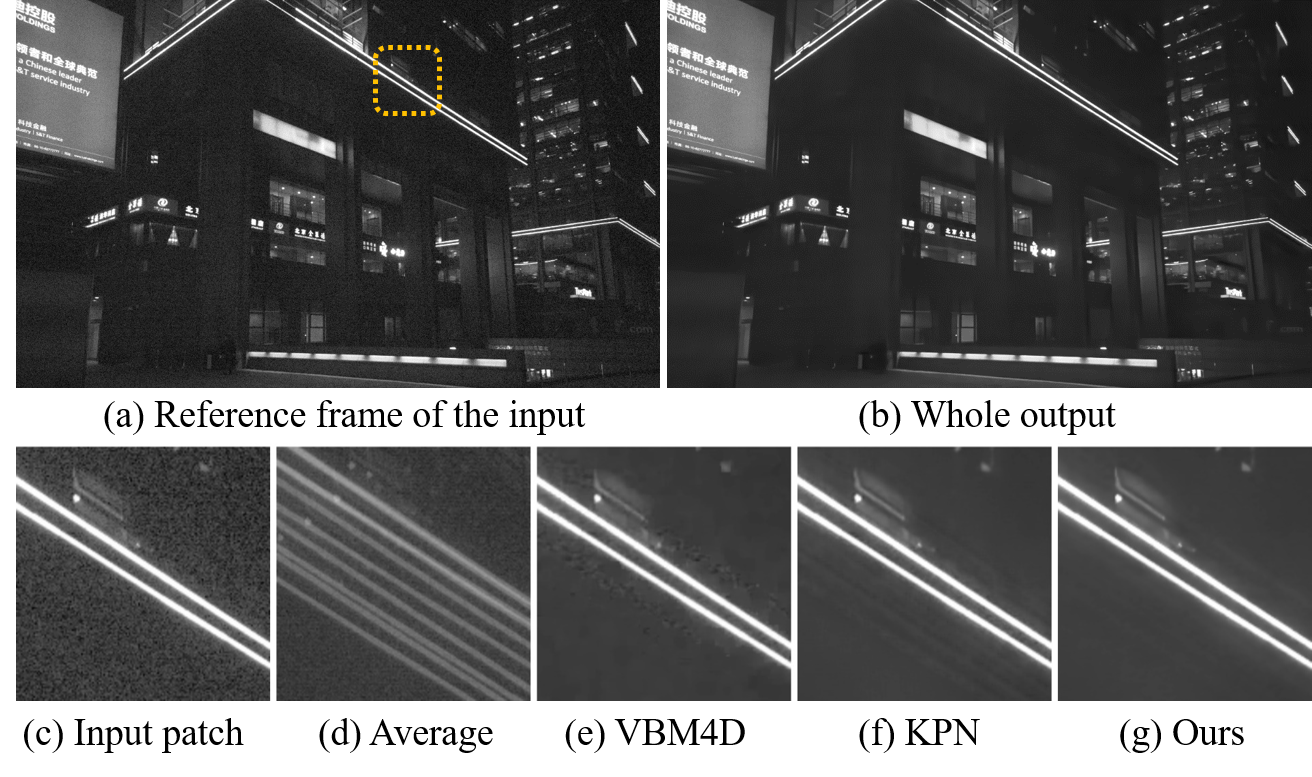} \\
		\end{tabular}
	\end{center}
	\vspace{-2mm}
	\caption{Video denoising results of a real captured sequence. (d) is generated by directly averaging the input frames. Note the ghosting artifacts around the glowing tubes by the KPN method in (f).
	}
	\label{fig:ghost}
	\vspace{-4mm}
\end{figure}

\subsection{Evaluation on the Proposed Dataset}
\label{sec: eval_our_data}
We evaluate the proposed algorithm against the state-of-the-art image and video denoising methods~\cite{KPN,vbm4d2012,bm3d2007,non-local-2005,dncnn} on the synthetic dataset at different noise levels. 
We conduct exhaustive hyper-parameter finetuning for the NLM~\cite{non-local-2005}, BM3D~\cite{bm3d2007} and VBM4D~\cite{vbm4d2012} methods including both blind and non-blind models, and choose the best results.
For fair comparisons, we train the KPN~\cite{KPN} and DnCNN~\cite{dncnn} methods on our datasets with the same settings.
While the KPN \cite{KPN} scheme is originally designed for multi-frame input, we 
adapt it to single image for more comprehensive evaluation by changing the network input.

As shown in Table~\ref{table:synthetic_result_image} and~\ref{table:synthetic_result_video}, the proposed algorithm achieves consistently better results on both single image and video denoising in terms of both PSNR and structural similarity (SSIM) in all the subsets with different noise levels.
Even our blind version model achieves competitive results whereas other methods rely on the oracle noise parameters to perform well.
Note that the KPN~\cite{KPN} learns convolution kernels for image and video denoising, where the irrelevant input pixels can negatively affect the filtering process and lead to inferior denoising results.
Table~\ref{table:synthetic_result_video} also shows the results by applying the PAN model on each frame separately and then fusing the outputs with weighted sum for video denoising (denoted as PAN-sep).
The proposed ST-PAN model can generate better results than PAN-sep owing to its capability of handling large motion.

\begin{table*}[]
	\centering
	\caption{Quantitative evaluation of video denoising on homoscedastic Gaussian noise. ``Tennis'', ``Old Town Cross'', ``Park Run'', and ``Stefan'' represent the 4 subsets of the video dataset~\cite{claus2019videnn}. 
	Since \cite{claus2019videnn} does not provide the SSIMs, we as well only compare the PSNRs in this table for clarity.}
	\begin{tabular}{l|c|c|c|c|c|c|c|c|c|c|c|c}
		\hline
		& \multicolumn{3}{c|}{\textit{Tennis}} & \multicolumn{3}{c|}{\textit{Old Town Cross}} & \multicolumn{3}{c|}{\textit{Park Run}} & \multicolumn{3}{c}{\textit{Stefan}} \\ \hline
		\multicolumn{1}{c|}{$\sigma$}  & 5       & 25      & 40      & 15          & 25         & 40        & 15        & 25       & 40      & 15       & 25      & 55      \\ \hline
		DnCNN~\cite{dncnn}    & 35.49   & 27.47   & 25.43   & 31.47      & 30.10      & 28.35     & 30.66    & 27.87    & 25.20   & 32.20   & 29.29   & 24.51   \\ \hline
		VBM4D~\cite{vbm4d2012}    & 34.64   & 29.72   & 27.49   & 32.40      & 31.21      & 29.57     & 29.99    & 27.90    & 25.84   & 29.90   & 27.87   & 23.83   \\ \hline
		ViDeNN~\cite{claus2019videnn}   & 35.51   & 29.97   & 28.00   & 32.15      & 30.91      & 29.41     & 31.04    & 28.44    & 25.97   & 32.06   & 29.23   & 24.63   \\ \hline
		ViDeNN-G~\cite{claus2019videnn} & 37.81   & 30.36   & 28.44   & 32.39      & 31.29      & 29.97     & 31.25    & 28.72    & 26.36   & 32.37   & 29.59   & 25.06   \\ \hline
		TOF~\cite{xue2019video} & 34.83   & 29.31   & 27.51   & 32.24      & 31.20      & 29.56     & 29.45    & 27.19    & 25.18   & 29.84   & 27.83   & 23.28   \\ \hline
		INN~\cite{kokkinos2019iterative} & 37.63 & 29.76 & 28.17 & 32.52 & 31.38 & 27.14 & 30.86 & 27.93 & 24.64 & 32.14 & 28.72 & 24.00 \\ \hline
		DVDnet~\cite{tassano2019dvdnet} &37.27& 30.47 & 28.25 & 32.54 & 31.72 & 29.93 & 31.17 & 28.70 & 26.12 & 31.73 & 29.04 & 24.09 \\ \hline
		VNLnet~\cite{davy2019non} & 38.25 & 30.58 & 28.09 &32.27&31.37& 30.35 & 31.21 & 28.76 & 26.15 & 32.39 & 29.55 & 24.55 \\ \hline
		KPN~\cite{KPN}  &   38.55 & 30.45 & 28.43 & 32.40 & 31.52 & 30.34 & 31.41 & 28.84 & 26.48 & 32.36 & 29.61 & 25.10     \\ \hline
		Ours     &  \bf{39.25} & \bf{30.73} & \bf{28.55} & \bf{33.19} & \bf{32.02} & \bf{30.62} & \bf{32.17} & \bf{29.47} & \bf{26.90} & \bf{32.59} & \bf{29.71} & \bf{25.22}  \\ \hline
	\end{tabular}
	\label{tab:more_vid}
\end{table*}

\begin{figure*}[t]
	\footnotesize
	\begin{center}
		\begin{tabular}{cccccc}
			\includegraphics[height = 0.11\linewidth]{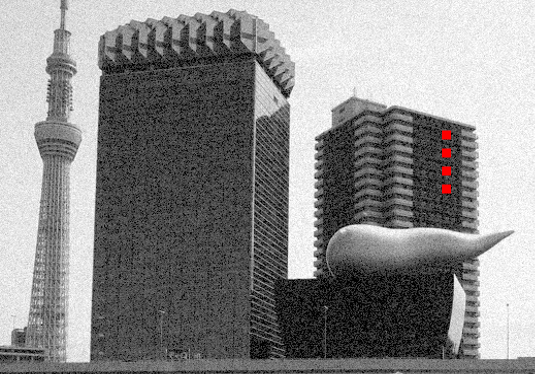} & \hspace{-4mm}
			\includegraphics[height = 0.11\linewidth]{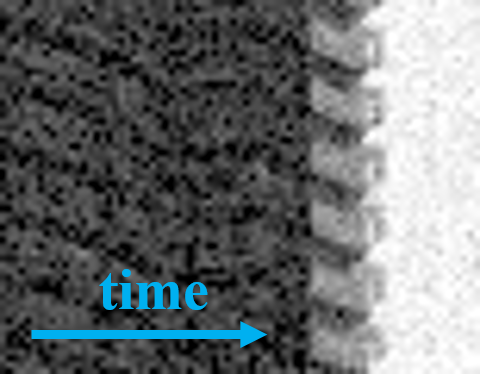}  & \hspace{-4mm}
			\includegraphics[height = 0.11\linewidth]{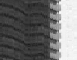} & \hspace{-4mm}
			\includegraphics[height = 0.11\linewidth]{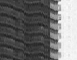} & \hspace{-4mm}
			\includegraphics[height = 0.11\linewidth]{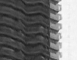} & \hspace{-4mm}
			\includegraphics[height = 0.11\linewidth]{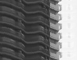} \\
			(a) Reference frame of input & \hspace{-4mm} (b) Input & \hspace{-4mm} (c) VBM4D~\cite{vbm4d2012} & \hspace{-4mm} (d) KPN~\cite{KPN} & \hspace{-4mm} (e) ST-PAN & \hspace{-4mm} (f) GT \\
		\end{tabular}
	\end{center}
	\caption{Temporal consistency of the proposed video denoising method. 
		We collect 1D samples over 60 frames from the red dashed line shown in (a), and concatenate these 1D samples into a 2D image to represent the temporal profiles of the videos.
		Specifically, (b)-(f) show the temporal profiles of the input, VBM4D~\cite{vbm4d2012}, KPN~\cite{KPN}, and our model, where the proposed ST-PAN model achieves better temporal consistency. 
	}
	\label{fig:temporal_effect}
\end{figure*}

Figure ~\ref{fig:synthetic_video_1} shows several image and video denoising results from the synthetic dataset. 
Conventional methods~\cite{bm3d2007,non-local-2005,vbm4d2012} with hand-crafted sampling and weighting strategies do not perform well and generate severe artifacts.
In particular, 
the VBM4D~\cite{vbm4d2012} method selects pixels using $L_2$ norm to measure patch similarities, which tends to generate oversmoothing results, as shown in Figure~\ref{fig:synthetic_video_1}(j).
On the other hand, directly synthesizing the results with deep CNNs~\cite{dncnn} can lead to denoised results with corrupted structures and fewer details (Figure~\ref{fig:synthetic_video_1}(e)).
Furthermore, the KPN~\cite{KPN} learns rigid kernels for video denoising, which do not deal with misalignments larger than $2$ pixels due to the limitation of rigid sampling.
When the misalignment is beyond this limit, the KPN model is likely to generate oversmoothed results (Figure~\ref{fig:synthetic_video_1}(k)) or ghosting artifacts around high-contrast boundaries, as shown in Figure~\ref{fig:ghost}.
In contrast, the proposed method learns the pixel aggregation process in a data-driven manner and achieves clearer results with fewer artifacts (Figure~\ref{fig:synthetic_video_1}(f), (l) and Figure~\ref{fig:ghost}(g)).

\subsection{Evaluation on Homoscedastic Gaussian Noise}
Whereas the noise in real world is mostly signal-dependent and heteroscedastic~\cite{healey1994radiometric,gonzalez2002digital,xu2019towards},
existing methods often evaluate their denoising algorithms on homoscedastic Gaussian noise~\cite{lefkimmiatis2017non,dncnn,Ploetz:2018:NNN,liu2018non,gu2019self,chen2019real,claus2019videnn,davy2019non,tassano2019dvdnet}.
For more comprehensive study, 
we evaluate the proposed PAN and ST-PAN models on image and video denoising datasets with homoscedastic Gaussian noise.
As shown in Table~\ref{tab:more_img},
our single image denoising model performs favorably against the baseline methods~\cite{nlnet,n3net,dncnn,nlrn,gu2019self,chen2019real,jia2019focnet} on the Set12 and BSD68 datasets~\cite{dncnn}. 
Since the original models of SGN~\cite{gu2019self} are not available, 
we train the SGN with the code provided by the authors following the settings of the original paper~\cite{gu2019self}.
Furthermore, the ST-PAN method achieves consistently better results than the state-of-the-art burst and video denoising approaches~\cite{vbm4d2012,claus2019videnn,kokkinos2019iterative,tassano2019dvdnet,davy2019non,xue2019video,KPN} on the dataset of \cite{claus2019videnn} under different noise levels (Table~\ref{tab:more_vid}).
\subsection{Temporal Consistency}
It is often desirable for the video denoising algorithms to generate temporally-coherent video frames.
In Figure~\ref{fig:temporal_effect}, we show some video denoising results for evaluating the temporal consistency of the proposed model.
Specifically, we collect 1D samples highlighted by the vertical red line (as shown in Figure~\ref{fig:temporal_effect}(a)) through 60 consecutive frames, and concatenate these 1D samples into a 2D image to represent the temporal profiles of the denoised videos.
Compared to the results of the baseline methods (Figure~\ref{fig:temporal_effect}(c) and (d)), the temporal profile of the proposed ST-PAN model (Figure~\ref{fig:temporal_effect}(e)) has smoother structures and fewer jittering artifacts, which indicates better temporal consistency of our model.
\subsection{Generalization to Real Inputs}
\begin{figure}[t]
	\footnotesize
	\begin{center}
		\begin{tabular}{c}
			\includegraphics[width = 0.93\linewidth]{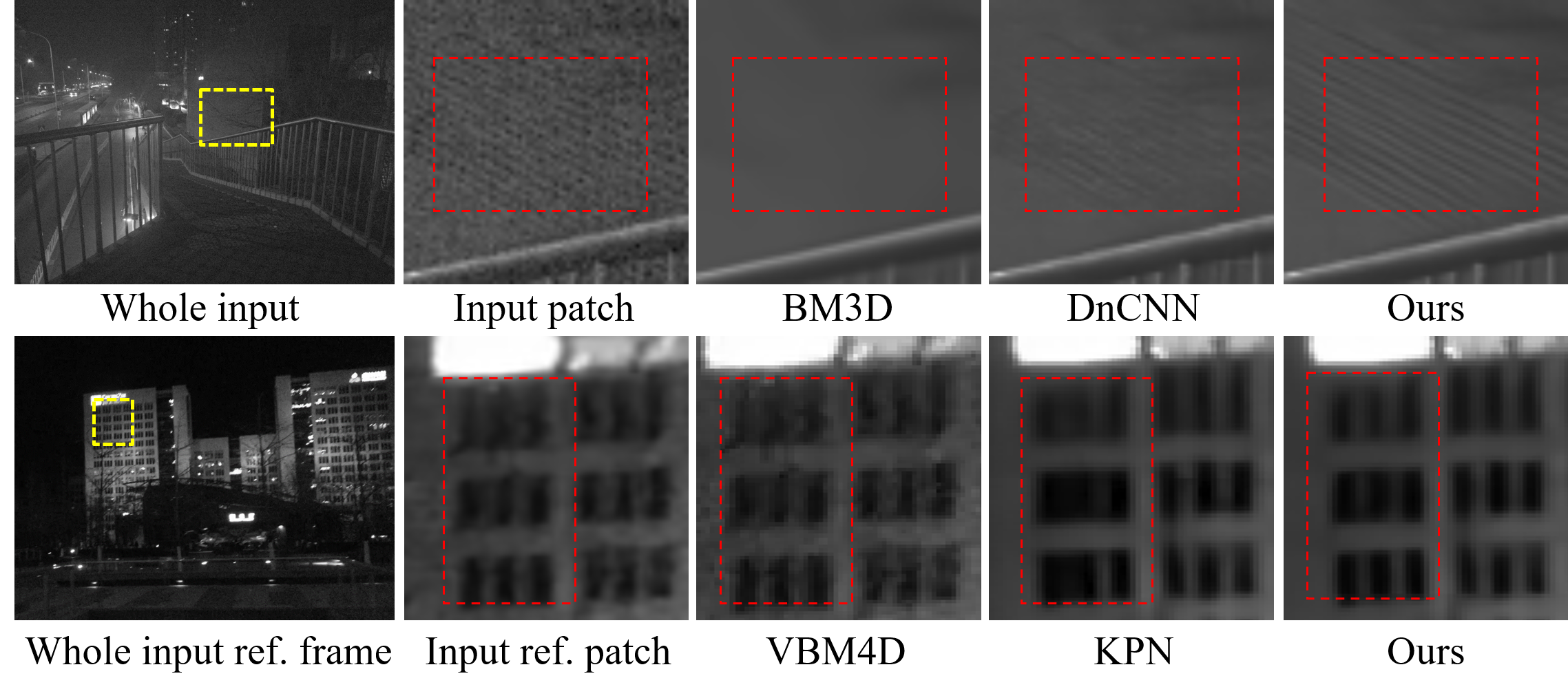} \\
		\end{tabular}
	\end{center}
	\vspace{-2mm}
	\caption{Results on real noisy image (first row) and video frame sequence (second row) captured by cellphones. ``ref.'' denotes the reference frame.
	}
	\label{fig:real_compare}
	\vspace{-5mm}
\end{figure}

We evaluate our method with state-of-the-art denoising approaches~\cite{bm3d2007,dncnn,KPN,vbm4d2012} on real images and video sequences captured by cellphones in Figure~\ref{fig:teaser} and \ref{fig:real_compare}.
While trained on synthetic data, 
our model is able to recover subtle edges from the real-captured noisy input and well handle misalignment from large motions.

\begin{figure*}[t]
	\footnotesize
	\begin{center}
		\begin{tabular}{c}
			\includegraphics[width = 0.8\linewidth]{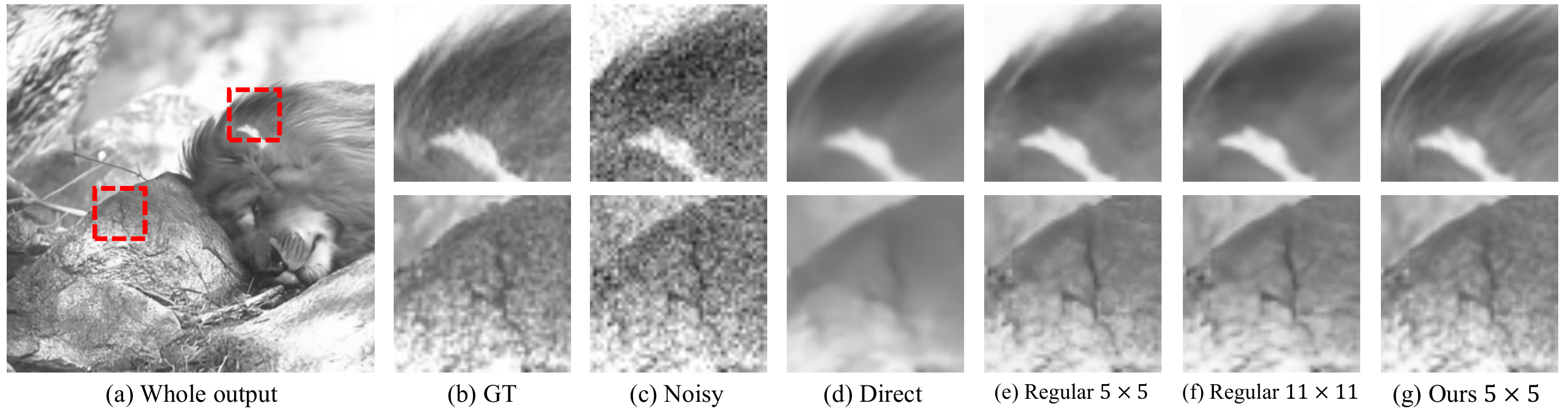} \\
		\end{tabular}
	\end{center}
	\vspace{-4mm}
	\caption{
		Denoised results by variants of the proposed model. 
		Pixel aggregation with learned sampling grid and weighting strategy achieves higher-quality result with fewer visual artifacts.
	}
	\label{fig:visual_compare}
	\vspace{-3mm}
\end{figure*}

\section{Discussion and Analysis}
\label{sec:discuss}
\subsection{Ablation Study}
In this section, we present ablation studies on different components of our algorithm for better analysis. 
We show the PSNR and SSIM for six variants of the proposed model in Table~\ref{table:ablation}, where ``our full model $3\times3\times$3'' is the default setting. 
First, the model ``direct'' uses the offset network in Figure~\ref{fig:3d_ker} to directly synthesize the denoised output, which cannot produce high-quality results.
This demonstrates the effectiveness of our model to learn the pixel aggregation for denoising. 
Second, to learn the spatially-variant weighting strategies, we use dynamic weights for the proposed pixel aggregation networks.
As shown in the second row of Table~\ref{table:ablation}, learning the model without dynamic weights significantly degrades the denoising performance.
On the third and fourth rows, we show the denoising results using rigid sampling grids with different sizes.
The result shows that learning the pixel sampling strategies is important for the denoising process and significantly improves the performance.
Furthermore, we concatenate the features of the offset network to predict the aggregation weights in Figure~\ref{fig:pipeline}.
The model without concatenating these features cannot exploit the deep offset network and only relies on the shallow structure of three convolution layers for weight prediction, which results in decreased performance as shown by the fifth row of Table~\ref{table:ablation}. 
In addition, the results on the sixth row show that the annealing term is important for training our model, and 
all components of our method are essential for denoising. 
Note that our learned sampling grid with size $3\times3\times3$ can sample pixels from a large receptive field (up to $\pm$15 pixels in our experiment), and further increasing the grid size of the ST-PAN only marginally improves the performance.
Thus, we choose a smaller sampling size as our default setting in this work.

\begin{table}[t]
	\begin{center}
		\footnotesize
		\caption{Ablation study on the synthetic dataset.
			\label{table:ablation}}
		\vspace{-3mm}
		\begin{tabular}{ l | cc | cc }
			\hline
			\multicolumn{1}{l|}{{\multirow{2}*{Algorithms}}} & 
			\multicolumn{2}{c|}{Low} & 
			\multicolumn{2}{c}{High} \\
			\cline{2-5}
			& PSNR & SSIM & PSNR & SSIM  \\ 
			\hline
			direct &
			35.45 & 0.9518 & 32.71 & 0.9200 \\
			fixed averaging weights &
			35.50 & 0.9449 & 32.60 & 0.9058 \\
			rigid sampling grid $3\times3\times3$ &
			36.02 & 0.9555 & 33.33 & 0.9256 \\
			rigid sampling grid $5\times5\times5$ &
			36.37 & 0.9592 & 33.73 & 0.9320 \\
			w/o concatenating offset features~~ &36.10&0.9590&33.46&0.9348 \\
			w/o regularization term & 
			36.16 & 0.9601 & 33.48 & 0.9341 \\ 
			our full model  $3\times3\times3$ & 
			\bf 36.91 & 0.9622 & 34.23 & 0.9366 \\
			our full model  $5\times5\times5$ &
			36.88 & \bf 0.9631 & \bf 34.25 & \bf 0.9379\\
			\hline
		\end{tabular}
		\vspace{-5mm}
	\end{center}
\end{table}

Except for the quantitative comparisons shown above, we present more detailed analysis as follows to illustrate why the different variants of our model do not perform well.  

{\flushleft \bf Direct synthesis.}
As introduced in Section~\ref{sec:intro}, 
the aggregation-based denoising process, including both pixel sampling and averaging, is usually spatially-variant and data-dependent. 
However, most CNNs use spatially-invariant and data-independent convolution kernels, and 
often require very deep structures to implicitly approximate the denoising process. 
Thus, direct synthesizing denoised outputs with CNNs is likely to result in local minimum solutions with over-smoothed results.
Similar findings have also been shown in \cite{niklaus2017video} for video frame interpolation.
In contrast, the proposed pixel aggregation model explicitly learn this spatially-variant filtering process, which can effectively exploit image structures to alleviate the aforementioned issues of direct synthesis.
In addition, our model directly aggregates input pixels, which constrains the output space and thus generate fewer artifacts in the denoised results.
A visual comparison between direct synthesis and our method is shown in Figure~\ref{fig:visual_compare}(d) and (g).

{\flushleft \bf Fixed averaging weights.}
As shown in \eqref{eq:motivation}, the principle of image and video denoising is to sample similar pixels around the one to be denoised and then take them as multiple observations of the input pixel for averaging.
Thus, a straightforward solution for image denoising is to only predict the sampling locations and average the predicted observations (\ie~sampled pixels) with the same weights for different pixel locations.
This is conceptually similar to the deformable convolution network~\cite{dai2017deformable}, which uses kernels with adaptive spatial shape and fixed parameters for object detection.
However, the sampled pixels from the noisy input usually do not obey the exact same distribution and thus should be adaptively weighted in the denoising process. 
For example, aggregation-based algorithms~\cite{tomasi1998bilateral,non-local-2005} exploit patch similarity to design weighting strategies for effective image denoising.
Similar to  these methods~\cite{tomasi1998bilateral,non-local-2005}, we learn content-aware averaging weights for pixel aggregation, which significantly improves the performance over fixed averaging weights in Table~\ref{table:ablation}.

\begin{figure}[t]
	\footnotesize
	\begin{center}
		\begin{tabular}{c}
			\includegraphics[width = 0.99\linewidth]{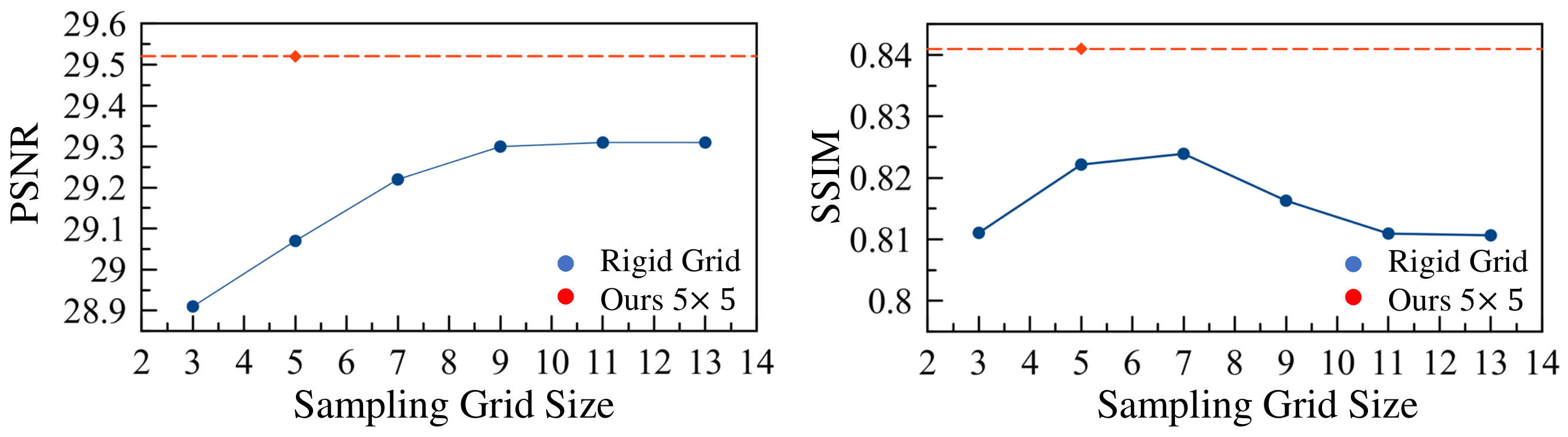} \\
		\end{tabular}
	\end{center}
	\vspace{-3mm}
	\caption{
		Comparing our method to the rigid sampling scheme with different grid sizes.
		The denoised results are obtained using the BSD68 dataset with $\sigma=25$.
	}
	\label{fig:compare_large_ker}
\end{figure}
\begin{figure*}[t]
	\footnotesize
	\begin{center}
		\begin{tabular}{c}
			\includegraphics[width = 0.8\linewidth]{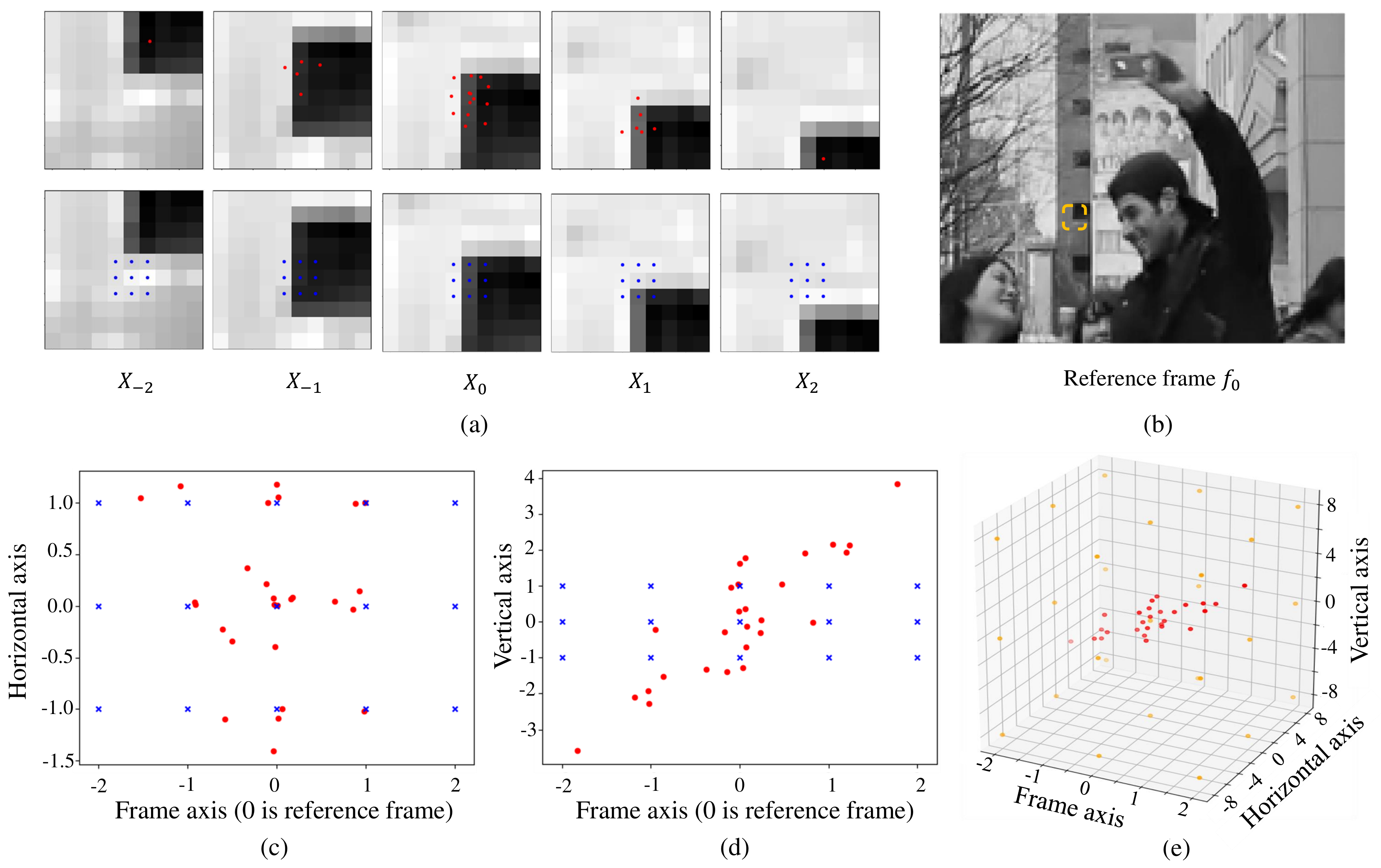} \\
		\end{tabular}
	\end{center}
\vspace{-3mm}
	\caption{
		Pixel aggregation process of the ST-PAN model for video input.
		The patch sequence $\{X_{-2},X_{-1},X_0,X_{1},X_2\}$ in (a) is cropped from the same spatial location of a sequence of consecutive video frames. We show the cropping location in the original reference frame (b) with an orange curve.
		The blue points in the bottom row of (a) denote \emph{five} rigid sampling grids with size $3\times3$, while the red points in the top row of (a) represent \emph{one} adaptive grid with size $3\times3\times3$.
		The center blue point in $X_0$ is the reference pixel for denoising.
		As the window in (a) is moving vertically, the sampling locations also moves vertically to trace the boundaries and search for more reliable pixels, which helps solve the misalignment issue.
		For better geometrical understanding, we show the 3D grids in the spatio-temporal space as the red points in (e).
		In addition, we respectively project the sampled pixels to different 2D planes as shown in (c) and (d).
		Note that higher coordinate in (d) indicates lower point in (a).
		With the frame index getting larger, the sampling locations distribute directionally in the Vertical axis of (d) while lying randomly in the Horizontal axis of (c), which is consistent with the motion trajectory of the window in (a).
	}
	\label{fig:learned_kernels}
	\vspace{-5mm}
\end{figure*}
\begin{figure}[t]
	\footnotesize
	\begin{center}
		\begin{tabular}{c}
			\hspace{-4mm}
			\includegraphics[width = 0.9\linewidth]{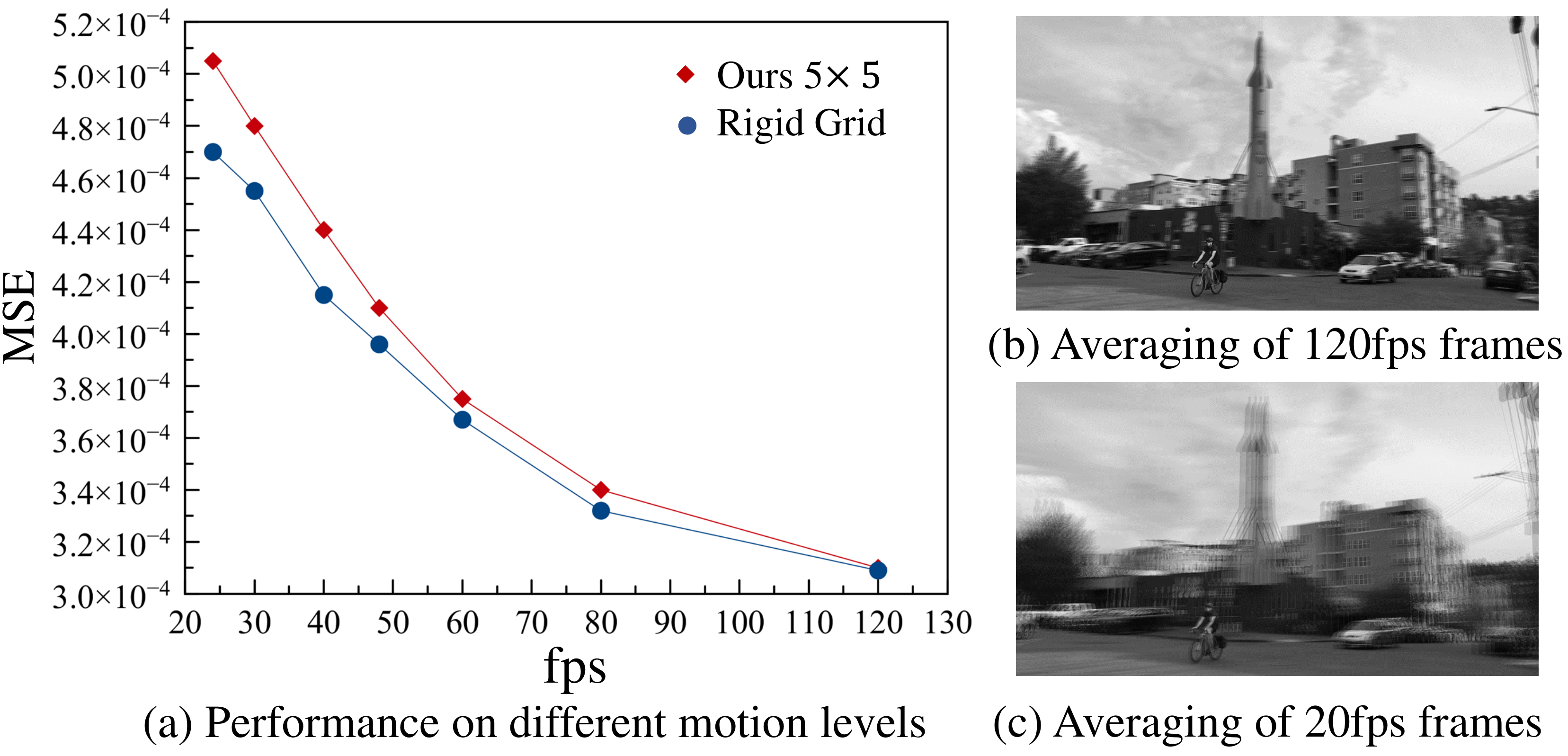} \\
		\end{tabular}
	\end{center}
\vspace{-3mm}
	\caption{Experimental results by the PAN-sep and ST-PAN models on different motion levels. Smaller frame rate at the fps-axis in (a) indicates larger motion. We visualize the motion difference by averaging the $120$fps and $24$fps input sequences in (b) and (c).
	}
	\label{fig:fps_test}
	\vspace{-3mm}
\end{figure}

\begin{figure}[t]
	\footnotesize
	\begin{center}
		\begin{tabular}{c}
			\includegraphics[width = 0.99\linewidth]{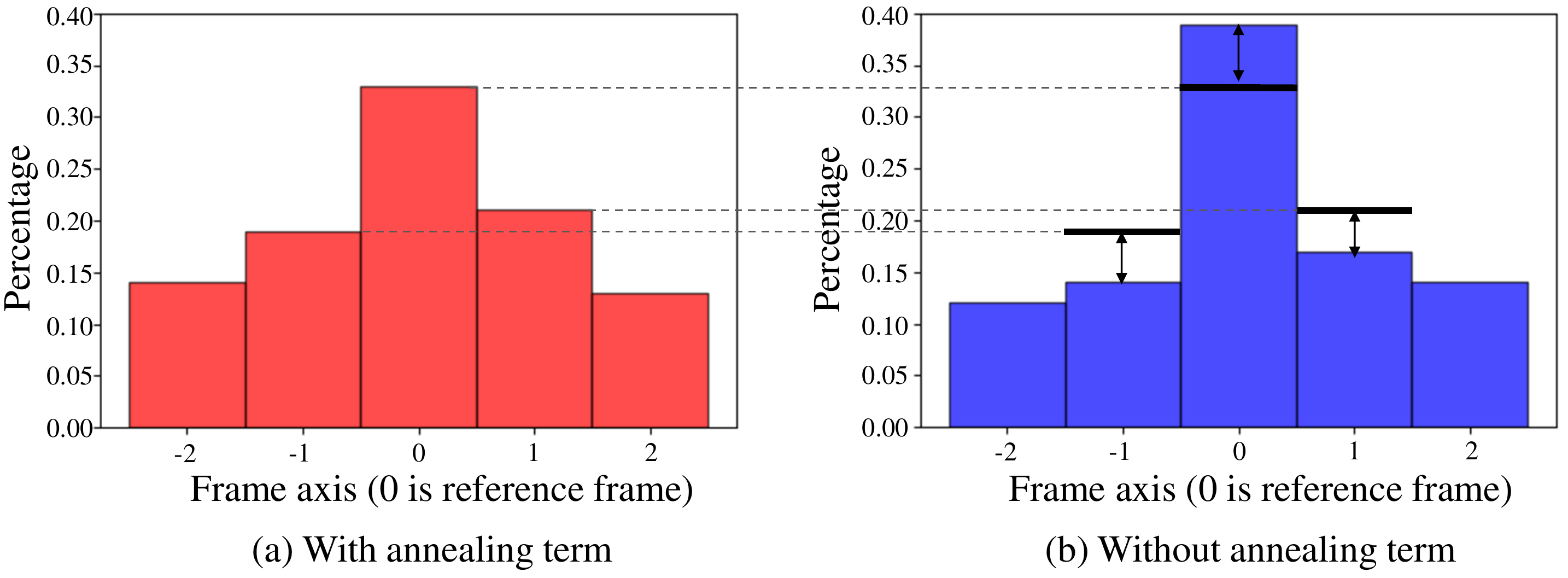} \\
		\end{tabular}
	\end{center}
\vspace{-3mm}
	\caption{Distributions of the sampling locations on the time dimension in the test dataset. (a) and (b) represent our models with and without using the annealing term. $x$- and $y$-axis denote the frame index and the percentage of pixels, respectively.}
	\label{fig:t_distribution}
	\vspace{-3mm}
\end{figure}

{\flushleft \bf Rigid sampling grid.}
Another straightforward alternative of the proposed method is to aggregate pixels from a rigid sampling grid. 
The influence of irrelevant sampling locations in the rigid grid can be reduced by the adaptive weighting model, which learns to give higher weights to more similar pixels. However, the rigid strategy can only sample pixels from a restricted receptive field, which hinders the model from utilizing more valid observations for denoising.
This can be understood by \eqref{eq:motivation}, where smaller $N$ leads to higher-variance estimates indicating worse denoised output (Figure~\ref{fig:visual_compare}(e)).
In contrast, the proposed algorithm can adapt to the image structures (as shown in Section~\ref{sec:vis}), and increase the receptive field without sampling more pixels. 
As such, it can exploit more useful observations (larger $N$ in \eqref{eq:motivation}) and reduce the variance of the estimated results, thereby leading to better denoising performance (Figure~\ref{fig:visual_compare}(g)). 

While our method addresses the issues of rigid sampling,
one potential question is whether the trivial solution, \ie,~simply enlarging the sampling grid to cover larger areas, can achieve similar results.
We evaluate the image denoising models using rigid sampling strategy with different grid sizes on the BSD68 dataset.
As shown in Figure~\ref{fig:compare_large_ker}, enlarging the sampling grid is only 
effective for smaller grid sizes.
In addition, the SSIM values decrease when the grid size becomes larger than $7$.
This is mainly due to the large amount of irrelevant sampling locations in the large rigid grids, and we empirically show that it is difficult to address this issue solely by the learned adaptive weights. 
We also present one denoised output using the large-size rigid sampling strategy in Figure~\ref{fig:visual_compare}(f), which demonstrates the significance of the proposed pixel aggregation network.
\subsection{Effectiveness of the Spatio-Temporal Pixel Aggregation}
As illustrated in Figure~\ref{fig:introduction}, the proposed ST-PAN model samples pixels across the spatial-temporal space for video denoising, and thus better handles large motion videos.
To further verify the effectiveness of the spatio-temporal sampling on large motion, we evaluate the PAN-sep and ST-PAN models under different motion levels.
Specifically, we sample $240$fps video clips with large motion from the Adobe240 dataset~\cite{dvd}.
We temporally downsample the high frame rate videos and obtain $7$ test subsets of different frame rates: $120$, $80$, $60$, $48$, $40$, $30$, and $24$fps, where each contains 180 input sequences.
Note that the sequences with different frame rates correspond to videos with different motion levels, and all the subsets use the same reference frames.
As shown in Figure~\ref{fig:fps_test}, the performance gap between the 2D and 3D strategies becomes larger as the frame rate decreases, which demonstrates the effectiveness of the spatial-temporal sampling on large motion.
We also notice that both methods achieve better results (smaller MSE) on videos with higher frame rates, which shows the importance of exploiting temporal information in video denoising.

\subsection{Effectiveness of the Regularization Term in \eqref{eq:regu}}
Figure~\ref{fig:t_distribution} shows the distributions of the sampling locations on the time dimension in the test dataset. 
Directly optimizing the $L_1$ loss without the annealing term in video denoising often leads to undesirable local minima where most the sampling locations are around the reference frame as shown in Figure~\ref{fig:t_distribution}(b).
By adding the regularization term in the training process, the network is forced to search more informative pixels across a larger temporal range, which helps alleviate the local minima issues  (Figure~\ref{fig:t_distribution}(a)).

\subsection{Visualization of the Pixel Aggregation Process}
\label{sec:vis}
\begin{figure}[t]
	\footnotesize
	\begin{center}
		\begin{tabular}{c}
			\includegraphics[width = 0.9\linewidth]{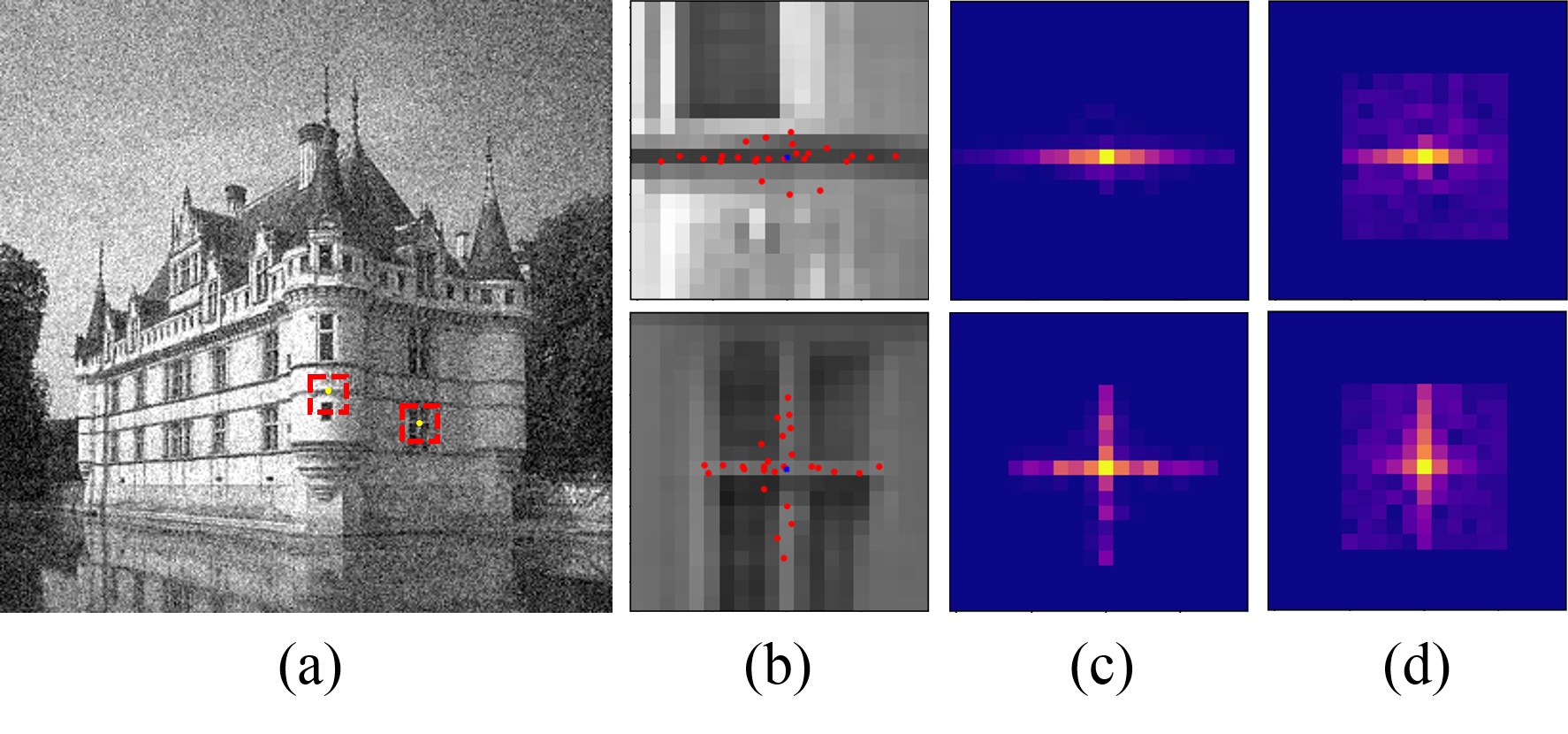} \\
		\end{tabular}
	\end{center}
	\vspace{-3mm}
	\caption{
		Visualization of the pixel aggregation process of PAN for single image input. 
		(a) is the noisy input.
		(b) represents the sampled pixels of PAN with grid size $5\times5$, and (c) shows the averaging weights of these pixels.
		We also show the weights of a large rigid sampling grid with size $11\times11$ in (d) for better understanding.
		Note that the PAN model achieves a large receptive field without increasing the grid size and reduces the influence of irrelevant pixels.
	}
	\label{fig:vis_2D_ker}
\end{figure}

For more intuitive understanding of the proposed algorithm, we visualize the denoising process of the PAN model in Figure~\ref{fig:vis_2D_ker}.
Our network exploits the structure information by sampling pixels along edges (Figure~\ref{fig:vis_2D_ker}(b) and (c)), and thereby reduces the interference of inappropriate samples for better image denoising performance.
Note that the predicted sampling locations in Figure~\ref{fig:vis_2D_ker}(b) are not always perfect mainly due to the noise of the input image.
However, the influence of the irrelevant sampling locations can be further reduced by the learned aggregation weights as shown in Figure~\ref{fig:vis_2D_ker}(c), where the out-of-edge pixels are given substially smaller weights for denoising. 
We also show a much larger rigid grid (size $11\times11$) in Figure~\ref{fig:vis_2D_ker}(d) to demonstrate why the straightforward strategy of increasing grid size does not work as well as our solution.

For video denoising, we show an example in Figure~\ref{fig:learned_kernels} to visualize the 3D sampling grid of the ST-PAN model. 
As shown in Figure~\ref{fig:learned_kernels}(a), the proposed ST-PAN model can trace the moving boundary along the motion direction and aggregate similar pixels from all the input frames.
The ability to sample both spatially and temporally is crucial for our method to deal with large motion and recover clean structures and details.

\subsection{Learned Location Shift for Different Noise Levels}
To provide further analysis of the learned sampling locations, we apply the proposed network to the same images with different noise levels and compute the average receptive field of the learned sampling grids in the images.
Table~\ref{table:shift_different_noise} shows that the network tends to search pixels from a wider area as the noise increases, which demonstrates that our model can automatically adjusts its sampling strategies to incorporate information from a larger receptive field to deal with more challenging inputs. 

\begin{table}[t]
	\begin{center}
		\footnotesize
		\caption{Average receptive field of the learned sampling grids under different noise levels on the proposed test dataset. The noise parameter ($\sigma_s, \sigma_r$) denotes the intensity of the shot noise and read noise. The proposed network tends to search pixels from a wider area as the noise intensity increases.
			\label{table:shift_different_noise}}
		\vspace{-3mm}
		\begin{tabular}{ l  ccc }
			\toprule
			Noise parameter & (2.5e-3, 1e-2) & (6.4e-3, 2e-2) & (1e-2, 5e-2)\\
			\midrule
			Average receptive field &6.8326& 7.294 & 7.7152 \\
			\bottomrule
		\end{tabular}
		\vspace{-5mm}
	\end{center}
\end{table}

\subsection{Running Speed}
The proposed method can process 0.27 megapixels in one second on a desktop with an Intel i5 CPU and a GTX 1060 GPU, which is close to the running speed of KPN (0.30 megapixels per second). 
Note that our model has a similar amount of parameters to KPN, and the additional time cost mainly comes from the trilinear sampler which can be potentially accelerated with parallel implementation.

\section{Conclusions}
In this work, we propose to learn the pixel aggregation process for image and video denoising with deep neural networks.
The proposed method adaptively samples pixels from the 2D or 3D input, 
handles misalignment caused by dynamic scenes, and enables large receptive fields while preserving details.
In addition, we present a regularization term for effectively training the proposed video denoising model.
Extensive experimental results demonstrate that our algorithm performs favorably against the state-of-the-art methods on both synthetic and real inputs.

While we use the inverse Gamma correction for synthesizing the training data, recent works have studied more realistic data generation in the raw domain~\cite{chen2019seeing,yue2020supervised,xu2019towards}.
Adapting the proposed network for raw data and exploring more realistic data generation pipelines will be interesting directions for future work.

\ifCLASSOPTIONcaptionsoff
  \newpage
\fi

{
\small
\bibliographystyle{IEEEtran}
\bibliography{egbib}
}

\end{document}